\documentclass[11pt]{article}
\usepackage[preprint]{acl}
\usepackage{times,latexsym}
\usepackage[T1]{fontenc}
\usepackage[utf8]{inputenc}
\usepackage{microtype,inconsolata}
\usepackage{graphicx,booktabs,array,colortbl,amsmath,placeins}
\usepackage{array,tabularx}
\usepackage{placeins}
\usepackage{xcolor}
\usepackage{tcolorbox}
\usepackage{tikz}
\usetikzlibrary{arrows.meta,positioning}
\tcbuselibrary{breakable}

\newtcolorbox{uncertaintycase}[2]{
  colback=#1!4!white,
  colframe=#1!55!black,
  colbacktitle=#1!12!white,
  coltitle=black,
  fonttitle=\bfseries,
  title={#2},
  title after break={#2 (continued)},
  boxrule=0.5pt,
  arc=1mm,
  left=5pt,
  right=5pt,
  top=5pt,
  bottom=5pt
}

\DeclareMathSizes{8.8}{9}{6}{5}
\DeclareMathSizes{8.4}{8}{6}{5}
\DeclareMathSizes{8.6}{9}{6}{5}
\DeclareMathSizes{8.2}{8}{6}{5}
\DeclareMathSizes{8.1}{8}{6}{5}

\definecolor{bestgray}{gray}{0.85}
\definecolor{secondgray}{gray}{0.94}
\newcommand{\best}[1]{\cellcolor{bestgray}\textcolor{black}{\textbf{#1}}}
\newcommand{\second}[1]{\cellcolor{secondgray}#1}
\title{LProp: Propagating Latent Uncertainty through Evidence-Conditioned Belief Revision in LLM Agents}
\author{
    Kaiwen Shi, Zheyuan Zhang \textsuperscript{$\dagger$}, 
    Weixiang Sun, Han Bao, Colby Nelson, Yanfang Ye 
    \\University of Notre Dame \\
    \textsuperscript{$\dagger$}Corresponding Author
    }
\begin{document}
\maketitle

\begin{abstract}
When LLM agents pass intermediate judgments downstream without uncertainty, later decisions may treat tentative assumptions as facts. Effective uncertainty propagation must preserve information for correction and update inherited beliefs as new evidence arrives. Yet scalar scores omit information needed for revision, while detailed descriptions and reflection incur additional costs. We introduce \textit{Latent Uncertainty}, a latent-space estimator that preserves uncertainty-relevant information without lengthy descriptions. Building on this, we propose \textit{LProp} (\textit{\underline{L}atent Uncertainty \underline{Prop}agation}), which revises inherited answer distributions using latent representations and new evidence without additional trajectory sampling. We evaluate both methods across multiple-choice, yes/no, open-ended question answering, and mathematical reasoning tasks. Our approach outperforms baselines on most metrics, while LProp achieves the lowest mean inference-token usage in all eight propagation settings. Matched feature ablations further demonstrate that inherited latent content improves error recovery beyond answer probabilities or model capacity alone. Our code is available \href{https://anonymous.4open.science/r/LProp-76C8/}{here}.\end{abstract}

\begin{figure}[t]
    \centering
    \includegraphics[width=\columnwidth]{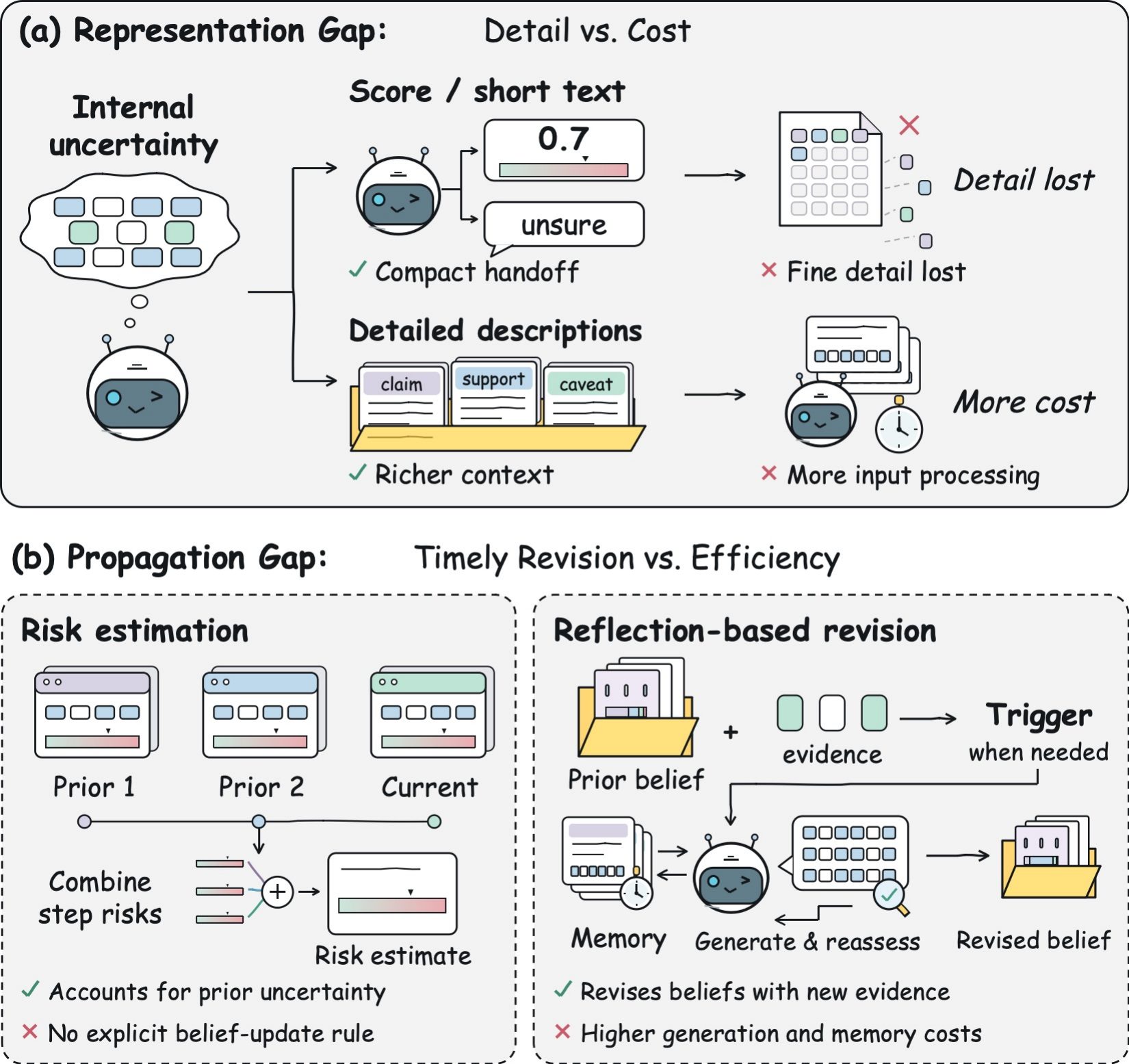}
    \caption{\textbf{Two gaps in uncertainty propagation.} (a) Compact representations lose uncertainty details, while detailed descriptions increase input cost. (b) Risk estimation alone does not specify how to revise beliefs, while reflection requires additional generation, motivating efficient and timely uncertainty updates.}
    \label{fig:uncertainty-gaps}
\end{figure}

\begin{figure*}[t]
\centering\includegraphics[width=\textwidth]{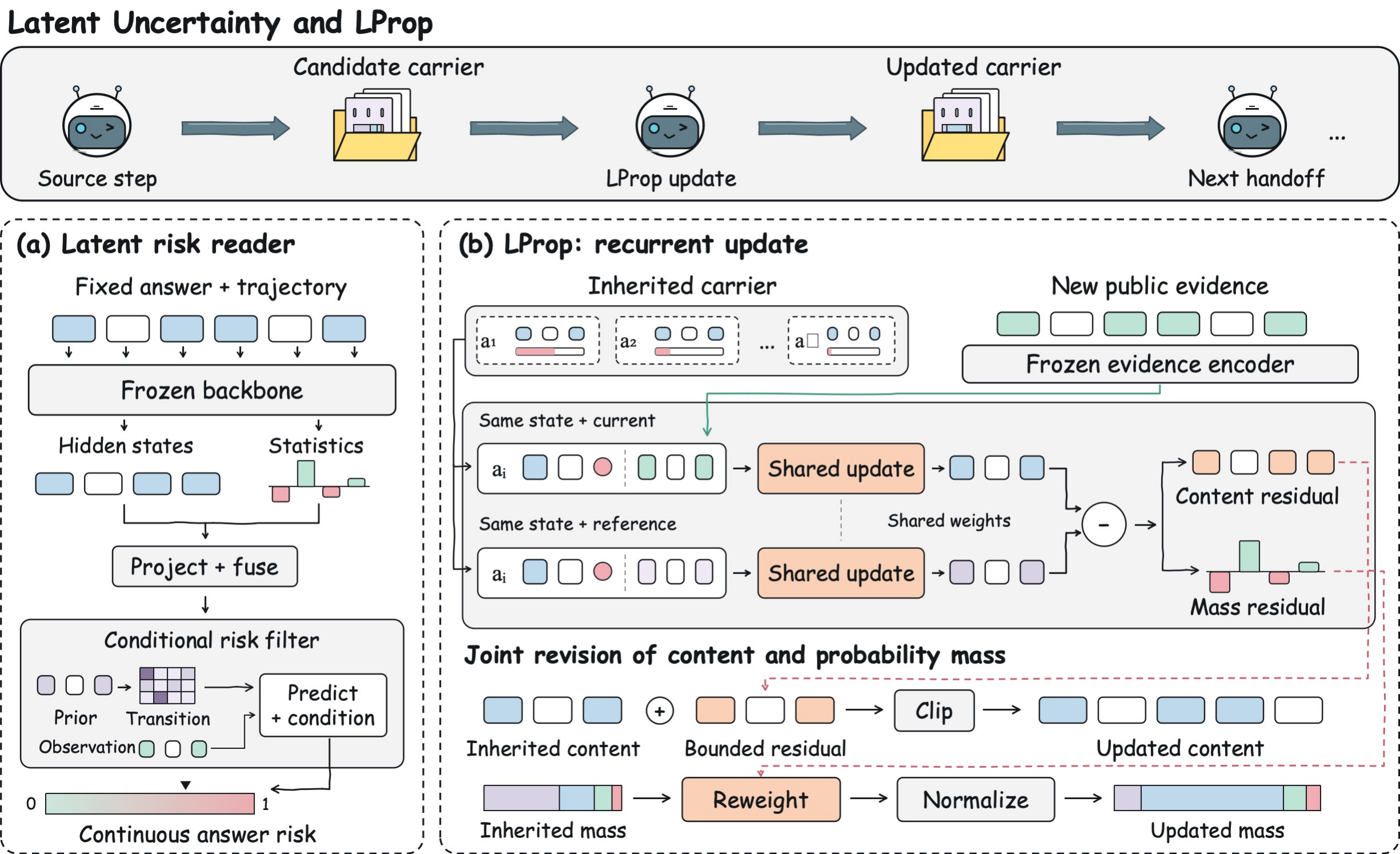}
\caption{\textbf{Overview of Latent Uncertainty and LProp.}
(a) The latent risk reader estimates answer-failure risk from hidden states and generation statistics, while an answer-indexed carrier retains candidate-specific content and support.
(b) LProp updates latent content and candidate probability mass with new evidence, enabling answer revision and uncertainty estimation across subsequent handoffs.}
\label{fig:methodology}
\end{figure*}

\section{Introduction}
\label{sec:introduction}

Nowadays, LLM agents are increasingly involved in people's work and daily lives, such as serving as personal assistants~\citep{openai2025agent, meta2026muse,ye2025llms4all}.
However, this growing reliance on LLM agents makes overconfidence consequential, as acting on unchecked assumptions can cause financial loss or data damage rather than merely produce an incorrect response~\citep{ruan2024toolemu}.
To address this risk, uncertainty can signal insufficient support for an agent's judgment and guide it to seek further information or assistance before committing~\citep{han-etal-2024-towards,shiuncertainty}.

\noindent\textbf{Preserving detail is costly.} Unlike single-response LLM tasks, agent tasks involve successive steps in which earlier results inform later decisions, a transfer we refer to as a \emph{handoff}~\citep{yao2023react,wu2024autogen}.
%
When a handoff preserves the result but omits its uncertainty, subsequent steps may treat tentative judgments as reliable premises, underestimating the uncertainty inherited from earlier decisions~\citep{duan2025uprop,zhao2025uncertainty,shi2026confidence}.
As a result, agents may proceed with unwarranted confidence and miss opportunities to verify assumptions inherited from earlier steps~\citep{zhang2026agenticuq}.
Therefore, uncertainty should be propagated throughout the agent's trajectory.
However, uncertainty propagation faces an \emph{information tax}: semantic entropy and verbalized confidence summarize uncertainty as scores~\citep{kuhn2023semantic,farquhar2024semanticentropy,tian-etal-2023-just}, but passing these scores alone does not convey the underlying alternatives needed for later revision (Figure~\ref{fig:uncertainty-gaps}a).
To reduce this loss, an agent can instead transmit detailed, structured descriptions of unresolved assumptions, alternative interpretations, and supporting evidence.
Yet retaining and extending these descriptions across steps creates a growing history that subsequent steps must process, increasing token cost and inference time~\citep{jiang-etal-2023-llmlingua}.
The challenge is therefore to preserve uncertainty-relevant information without continually expanding the communication and processing burden.

\noindent\textbf{Revision can be costly and unreliable.} Moreover, uncertainty propagation must determine not only how much uncertainty is inherited, but also how it should change with new evidence.
Existing methods account for inherited uncertainty through decomposition or situation-aware weighting~\citep{duan2025uprop,zhao2025uncertainty}.
However, these approaches primarily assess propagated uncertainty rather than actively revising inherited beliefs as new evidence arrives.
To support such revision, AUQ retains confidence and explanations in memory and triggers corrective reflection before execution~\citep{zhang2026agenticuq}.
Yet accumulated textual memory may introduce additional uncertainty as stored information becomes stale or inconsistent, while memory-based reflection incurs additional context-processing and generation costs~\citep{shinn2023reflexion,zhang2026agenticuq}.
The challenge is therefore to develop a reliable and efficient evidence-conditioned update that revises inherited beliefs as new evidence arrives, without relying on potentially stale memory or costly reflection (Figure~\ref{fig:uncertainty-gaps}b).

To address these challenges, we introduce \textit{Latent Uncertainty}, a latent-space uncertainty estimator that preserves uncertainty-relevant information across steps without lengthy textual descriptions. Building on this, \textit{LProp} learns an efficient, evidence-conditioned update that revises inherited uncertainty and adjusts its influence on subsequent decisions.
We evaluate our approach on multiple-choice, yes/no, open-ended question-answering, and mathematical reasoning tasks.
The results show consistently strong performance across task types, with improvements over the compared baselines on most evaluation metrics. Meanwhile, LProp also achieves the lowest mean inference-token usage in every evaluated propagation setting.
Our contributions are as follows.
\begin{itemize}
    \setlength{\itemsep}{2pt}
    \setlength{\parskip}{0pt}
    \item We introduce \textit{Latent Uncertainty}, a latent-space uncertainty estimator, and validate its ability to predict answer errors, providing an empirical foundation for uncertainty propagation through latent representations.
    \item Building on this foundation, we propose \textit{LProp}, an efficient propagation method that preserves inherited latent information and updates beliefs as new evidence arrives. Experiments demonstrate substantial error recovery across diverse tasks, with the lowest mean inference-token usage in all eight propagation settings.
    \item Through matched feature ablations, we demonstrate that inherited latent content improves downstream error recovery beyond answer probabilities or model capacity alone, establishing its contribution to evidence-conditioned belief revision.
\end{itemize}

\section{Related Work}

\subsection{Uncertainty Carrier}
The representation used to carry uncertainty determines what information remains available for downstream decisions. Existing methods make uncertainty explicit through probability-derived scores or verbalized confidence~\cite{malinin2020uncertainty, tian-etal-2023-just}. Token likelihoods and predictive entropy provide probability-based estimates, while semantic entropy groups sampled answers by semantic equivalence to quantify uncertainty over meanings~\citep{malinin2020uncertainty, kuhn2023semantic,farquhar2024semanticentropy}. Verbalized confidence instead asks models to report their confidence in text~\citep{tian-etal-2023-just,shi2026sage}. These outputs communicate the degree of uncertainty, but passing a score or brief confidence statement alone does not necessarily preserve the context needed to determine what to verify or revise. 

\subsection{Latent Information and Communication}
\label{sec:related-latent}
A separate line of work explores communication through hidden representations to preserve information beyond explicit outputs. CIPHER~\citep{phamCIPHER2024} passes probability-weighted embeddings between agents, Interlat~\citep{du-etal-2026-enabling} transmits hidden states with learned compression, and LatentMAS~\citep{zou2025latentmas} maintains shared latent working memory. Related work also explores extracting uncertainty-related signals from hidden representations. Azaria and Mitchell~\citep{azaria2023internal} train classifiers on hidden activations to predict statement truthfulness, while Semantic Entropy Probes~\citep{kossen2024semanticentropyprobesrobust} estimate semantic entropy from a single generation's hidden states. These findings motivate our latent-space uncertainty estimator, while our approach further introduces an answer-indexed carrier to preserve candidate-specific information for downstream evidence-conditioned revision.

\subsection{Uncertainty Propagation}
In multistep decision-making, uncertainty in earlier results can affect later decisions, making local confidence alone insufficient. SAUP~\citep{zhao2025uncertainty} aggregates step uncertainties using situation-aware weights based on reasoning drift and observation alignment. To capture isolated failure signals that averaging can dilute, TRACER~\citep{tayebati2026tracer} combines local uncertainty with repetition and coherence indicators, emphasizing high-risk episodes. While these methods aggregate observed risk, UProp~\citep{duan2025uprop} samples trajectories and alternative decisions to distinguish intrinsic from inherited uncertainty. Risk estimation alone, however, does not specify how to revise beliefs with new evidence. AUQ~\citep{zhang2026agenticuq} supports such revision by retaining confidence and explanations in memory and triggering reflection before execution. However, stored assessments may become stale or inconsistent, introducing further uncertainty, while memory-based reflection adds context-processing and generation costs. These limitations motivate us to develop an efficient, evidence-conditioned update that revises inherited beliefs in latent space without generating textual reflections.

\section{Methodology}
\label{sec:methodology}
An uncertainty score can flag a need for verification, but revising an answer also requires information about competing hypotheses. Latent Uncertainty therefore combines a reader that estimates the risk of a fixed answer with a separate carrier that retains candidate-specific content and support. LProp uses subsequent evidence to revise this support through a shared answer-distribution interface. Figure~\ref{fig:methodology} illustrates the two components. The backbone remains frozen throughout.

\subsection{Latent Uncertainty Carrier}
\label{sec:latent-uncertainty}

\paragraph{Reading answer risk.}
Given a trajectory with hidden states $v_{1:T}$ and a fixed answer $\hat a$, we construct step features $x_t$ from projected states and observable generation statistics. Projection and feature normalization are fitted on training data and frozen before evaluation. A learned map of $x_t$ produces a row-normalized transition matrix $T_t$ and a positive observation vector $o_t$. Starting from a learned prior $q_0$, a conditional filter integrates these observations and reads out answer-failure risk:
\begin{align}
q_t &= \mathcal{N}\!\left[(q_{t-1}T_t)\odot o_t\right],
\label{eq:risk-filter}\\
\widehat r &= \sum_{j=1}^{R}q_T(j)c_j.
\label{eq:risk-readout}
\end{align}
Here $q_t$ is a distribution over $R$ latent risk components, $\mathcal{N}$ normalizes a nonnegative vector to sum to one, and $\odot$ denotes elementwise multiplication. The transition carries earlier observations forward, while the observation vector incorporates the current step. Fixed, ordered anchors $c_j\in[0,1]$ map the component probabilities to a continuous risk estimate. The reader is trained with binary cross-entropy on labels indicating whether $\hat a$ is incorrect, without changing the answer. This supervision concerns answer correctness and does not constrain the answer space. Additional reliability observations can refine the same estimate; feature construction and risk prediction are detailed in Appendix~\ref{app:risk-reader}.

\paragraph{Retaining answer-specific information.}
Risk estimation alone does not retain the alternatives needed for revision. We therefore construct a carrier over a support $\mathcal{A}_0=\{a_i\}_{i=1}^{K}$ available at the source:
\begin{equation}
\mathcal{C}_0=\{(a_i,h_i^0,w_i^0)\}_{i=1}^{K},
\qquad \sum_{i=1}^{K}w_i^0=1,
\label{eq:latent-carrier}
\end{equation}
where $a_i$ identifies a hypothesis, $h_i^0$ retains its projected source content of dimension $d$, and $w_i^0$ records its relative support. Initial weights are derived from source answer scores, with a source readout adjusting them where enabled. The $R$ risk components describe the reliability of a fixed answer, while the $K$ carrier entries describe competing answer hypotheses; the two distributions serve distinct purposes.

Candidate construction uses neither future evidence nor gold answers. Candidate identities remain fixed during propagation, and an aggregate residual category can retain probability mass outside the explicitly represented answers. The receiver obtains the carrier, the question, and later public evidence without the source's raw private context. The carrier's numerical representation contains $O(Kd)$ values, retaining candidate-specific information for revision over the source support. Construction and initialization details appear in Appendix~\ref{app:carrier-details}.

\subsection{LProp: Uncertainty Propagation through Latent Carrier}
\label{sec:lprop}

\paragraph{Evidence-conditioned support revision.}
At handoff $t$, LProp revises a distribution over the retained support. Let $b_t$ denote the inherited distribution supplied to the update. Pooling parameterizations additionally use $p_t^{\mathrm{pub}}$, the distribution obtained from public evidence over the same support. The inherited distribution and candidate content may be carried forward from the preceding revision or retained from the source. We express support revision in the common form
\begin{equation}
\begin{aligned}
\widetilde w_i^t &=(1-\alpha_t)b_t(i)
+\alpha_t p_t^{\mathrm{pub}}(i),\\
w_i^t &\propto \widetilde w_i^t\exp(\rho_i^t),
\end{aligned}
\label{eq:public-pooling}
\end{equation}
where $\alpha_t\in[0,1]$ controls reliance on public-evidence probabilities and $\rho_i^t$ is a candidate-specific evidence residual. The mixture determines how inherited and public support are combined, while the residual adjusts individual hypotheses using retained content and current evidence. Normalization covers the entire support, including any residual category.

This form admits both direct residual revision, with $\alpha_t=0$, and adaptive probability pooling. In the latter, a gate evaluates candidate content, inherited mass, distribution entropy, and evidence-change features; averaging its values under $b_t$ yields $\alpha_t$. These parameterizations share the carrier and revised-distribution interface, while differing in how they incorporate evidence. Propagation updates and task-specific parameterizations are summarized in Appendix~\ref{app:propagation-update}.

\paragraph{Revising latent content and support.}
Candidate content can remain anchored to the source or evolve as evidence is incorporated. For a recurrent reference-difference update, let $z_i^{t-1}$ collect the candidate's latent content, answer identity, and inherited support features. A shared function $F_\theta$ predicts content and log-mass changes under current evidence $e_t$ and reference evidence $e_t^{\mathrm{ref}}$:
\begin{equation}
\begin{split}
(\Delta h_i^t,\Delta \ell_i^t)
={}&F_\theta(z_i^{t-1},e_t)\\
&-F_\theta(z_i^{t-1},e_t^{\mathrm{ref}}).
\end{split}
\label{eq:evidence-difference}
\end{equation}
Holding the inherited state fixed in both evaluations measures the function's change in response to evidence relative to the chosen reference. The reference may encode the question alone or the preceding evidence state. The resulting increments jointly revise content and support:
\begin{equation}
\begin{aligned}
h_i^t &= \operatorname{clip}\!\left(h_i^{t-1}
+\gamma\tanh(\Delta h_i^t),-M,M\right),\\
w_i^t &\propto w_i^{t-1}
\exp\!\left(\Delta\ell_i^t+\nu_i^t\right),
\end{aligned}
\label{eq:joint-update}
\end{equation}
where $\gamma>0$ scales the bounded content residual, $M>0$ bounds the updated coordinates, and $\nu_i^t$ is an optional correction derived from changes in the public model's relative answer support. This mass update instantiates Equation~\ref{eq:public-pooling} with $b_t=w^{t-1}$, $\alpha_t=0$, and $\rho_i^t=\Delta\ell_i^t+\nu_i^t$. Revised content and masses then enter the next recurrent state. Task-specific reference choices and update parameterizations are summarized in Appendix~\ref{app:propagation-update}.

\begin{figure*}[t]
\centering\includegraphics[width=0.9\textwidth]{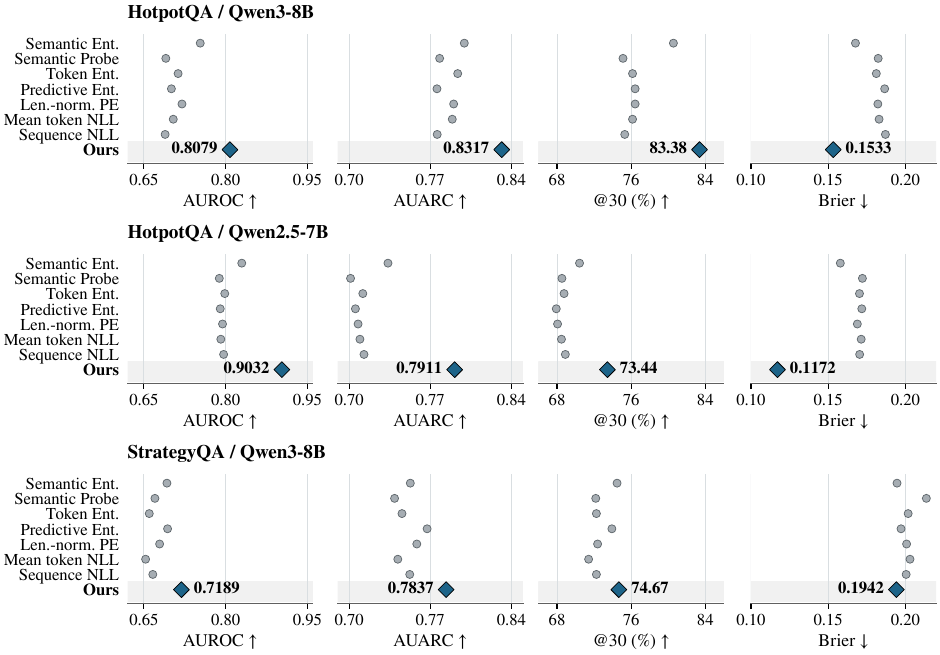}
\caption{\textbf{Latent uncertainty as an error-risk estimator.} Reader metrics are averaged over five training seeds. Rows show settings; columns show metrics. Gray circles: baselines; blue diamonds: Ours. Axes are zoomed and matched within each metric. Values and calibration details appear in Table~\ref{tab:estimation} and Appendix~\ref{sec:estimator-calibration}.}
\label{fig:estimation}
\end{figure*}

\paragraph{Learning and readout.}
Propagation is supervised at available handoff views using
\begin{equation}
\mathcal{L}_{\mathrm{core}}
=-\log w^t(y)+\lambda_{\mathrm B}B(w^t,y),
\label{eq:propagation-loss}
\end{equation}
where $y$ identifies the target entry in the retained support and $B$ is a Brier penalty. Views are grouped by question to prevent unequal view counts from dominating training, and model choices use development data. At inference, candidate probabilities are averaged across independently trained heads before selecting the admissible answer with the largest mass. For a valid selected answer, uncertainty is one minus its probability in the full distribution, a readout distinct from the reader's fixed-answer risk estimate $\widehat r$. Target construction, training, and inference details are given in Appendix~\ref{app:training-readout}. LProp encodes new public evidence and revises inherited support through the learned update, avoiding textual reflections and repeated trajectory sampling at each handoff.

\section{Experiments}
\label{sec:experiments}

We evaluate two requirements separately. The first experiment tests whether the Latent Uncertainty Estimator identifies errors in fixed agent answers. The second tests whether LProp improves answer quality and uncertainty assessment after evidence handoffs, and examines the token cost required to obtain these improvements.

\subsection{Latent Uncertainty Estimator}
\label{sec:estimator-experiments}

\subsubsection{Experimental Setting}
\textbf{Benchmarks.}
We evaluate the Latent Uncertainty Estimator on HotpotQA~\citep{yang-etal-2018-hotpotqa} with Qwen3-8B~\citep{yang2025qwen3technicalreport} and Qwen2.5-7B~\citep{qwen2025qwen25technicalreport} and on StrategyQA~\citep{geva-etal-2021-aristotle} with Qwen3-8B, using the released 1,024-question evaluations with 512 training and 128 development questions per setting.

\noindent\textbf{Compared methods.}
We compare against Token Entropy~\citep{fomicheva2020unsupervised}, Mean token NLL~\cite{fomicheva2020unsupervised}, Sequence NLL~\citep{malinin2020uncertainty}, Predictive Entropy~\citep{malinin2020uncertainty}, Length-normalized Predictive Entropy~\citep{malinin2020uncertainty}, Semantic Entropy~\citep{kuhn2023semantic}, and Semantic Entropy Probe~\citep{kossen2024semanticentropyprobesrobust}. The probe predicts semantic entropy and uses this prediction as its uncertainty score; its remaining evaluation settings match those of Semantic Entropy. Detailed evaluation protocols and baseline configurations appear in Appendix~\ref{sec:estimator-protocol}.

\noindent\textbf{Metrics.}
We report AUROC for error discrimination, AUARC for retained-answer accuracy across coverage levels, @30 for accuracy after rejecting the highest-risk 30\% of answers, and Brier for the squared error of predicted failure probabilities. Higher values are better for the first three metrics, while lower Brier scores are better.

\subsubsection{Results}
\textbf{Error discrimination.}
Figure~\ref{fig:estimation} shows that the Latent Uncertainty Estimator achieves the best point estimates for AUROC, AUARC, @30, and calibrated Brier in every evaluated setting. Because all estimators score the same fixed answers, these gains reflect better uncertainty estimation rather than changes in answer quality. The AUROC improvements are accompanied by higher AUARC and @30, indicating that better error discrimination also translates into more reliable retained answers across coverage levels.

\noindent\textbf{Probability estimates.}
Lower calibrated Brier point estimates complement the ranking improvements by assessing the quality of predicted failure probabilities. The clearest gains occur on HotpotQA. Together, these results support latent representations as a useful basis for assessing answer reliability. Appendix~\ref{app:estimator-results} provides the detailed comparisons.

\begin{table*}[t]
\centering
\fontsize{8.8}{10.4}\selectfont
\setlength{\tabcolsep}{2.1pt}
\renewcommand{\arraystretch}{1.13}
\begin{tabularx}{\textwidth}{@{}l*{12}{>{\centering\arraybackslash}X}@{}}
\toprule
 & \multicolumn{6}{c}{\textbf{Qwen3-8B}}
 & \multicolumn{6}{c}{\textbf{Mistral-7B-Instruct-v0.3}} \\
\cmidrule(lr){2-7}\cmidrule(lr){8-13}
Method
& \shortstack[c]{F1$\uparrow$}
& \shortstack[c]{AUROC$\uparrow$}
& \shortstack[c]{AUARC$\uparrow$}
& \shortstack[c]{@20$\uparrow$}
& \shortstack[c]{@30$\uparrow$}
& \shortstack[c]{@50$\uparrow$}
& \shortstack[c]{F1$\uparrow$}
& \shortstack[c]{AUROC$\uparrow$}
& \shortstack[c]{AUARC$\uparrow$}
& \shortstack[c]{@20$\uparrow$}
& \shortstack[c]{@30$\uparrow$}
& \shortstack[c]{@50$\uparrow$} \\
\midrule
\multicolumn{13}{@{}l}{\textbf{BoolQ}} \\
SAUP & 0.7630 & 0.5902 & 0.8347 & 0.8000 & 0.8114 & 0.8340 & 0.6990 & \second{0.5794} & \second{0.7136} & 0.7225 & 0.7343 & 0.7540 \\
TRACER & 0.7570 & 0.5076 & 0.7927 & 0.7838 & 0.7886 & 0.7920 & 0.6790 & 0.4730 & 0.6575 & 0.6912 & 0.6857 & 0.6820 \\
AUQ & 0.7480 & \second{0.7038} & \second{0.8520} & 0.7989 & \second{0.8164} & \second{0.8489} & 0.6650 & 0.4191 & 0.5383 & 0.6687 & 0.6588 & 0.6272 \\
UProp & \second{0.7870} & 0.5581 & 0.7976 & \second{0.8162} & 0.8143 & 0.8080 & \second{0.7190} & 0.5757 & 0.7066 & \second{0.7349} & \second{0.7414} & \second{0.7560} \\
\textbf{LProp} & \best{\underline{0.8310}} & \best{\underline{0.7961}} & \best{\underline{0.9407}} & \best{\underline{0.8950}} & \best{\underline{0.9186}} & \best{\underline{0.9460}} & \best{\underline{0.8010}} & \best{\underline{0.7656}} & \best{\underline{0.9178}} & \best{\underline{0.8600}} & \best{\underline{0.8829}} & \best{\underline{0.9200}} \\
\midrule
\multicolumn{13}{@{}l}{\textbf{StrategyQA}} \\
SAUP & 0.6417 & 0.5068 & 0.6528 & 0.6406 & 0.6399 & 0.6583 & 0.5875 & 0.5068 & 0.5497 & 0.5938 & 0.6101 & 0.5917 \\
TRACER & 0.6563 & 0.5396 & 0.6533 & 0.6510 & 0.6548 & 0.6875 & \second{0.5924} & \second{0.5344} & \second{0.5653} & \second{0.6146} & \second{0.6190} & 0.6083 \\
AUQ & \second{0.6750} & \second{0.6005} & \second{0.7209} & \second{0.7253} & \second{0.7195} & \second{0.7201} & 0.5667 & 0.3979 & 0.4300 & 0.5632 & 0.5476 & 0.5063 \\
UProp & 0.6438 & 0.4619 & 0.6037 & 0.6484 & 0.6369 & 0.6167 & 0.5896 & 0.5191 & 0.5487 & 0.6042 & 0.6071 & \second{0.6208} \\
\textbf{LProp} & \best{\underline{0.7250}} & \best{\underline{0.6865}} & \best{\underline{0.8372}} & \best{\underline{0.7682}} & \best{\underline{0.7946}} & \best{\underline{0.8417}} & \best{\underline{0.6500}} & \best{\underline{0.6160}} & \best{\underline{0.7415}} & \best{\underline{0.6797}} & \best{\underline{0.6905}} & \best{\underline{0.7375}} \\
\midrule
\multicolumn{13}{@{}l}{\textbf{MMLU}} \\
SAUP & 0.8854 & 0.6481 & \second{0.9263} & 0.9127 & 0.9182 & 0.9236 & 0.4586 & 0.3696 & 0.2967 & 0.4643 & 0.4455 & 0.3885 \\
TRACER & 0.8885 & 0.5006 & 0.8946 & 0.8770 & 0.8909 & 0.8854 & 0.4554 & 0.3099 & 0.2749 & 0.4246 & 0.4045 & 0.3376 \\
AUQ & \second{0.8917} & \second{0.6669} & 0.8882 & \second{0.9339} & \second{0.9409} & \second{0.9326} & \second{0.4682} & 0.3651 & 0.3036 & 0.4718 & 0.4504 & 0.3826 \\
UProp & 0.8885 & 0.6425 & 0.9186 & 0.9206 & 0.9318 & 0.9299 & 0.4586 & \second{0.4496} & \second{0.3329} & \second{0.4960} & \second{0.4955} & \second{0.4459} \\
\textbf{LProp} & \best{\underline{0.8949}} & \best{\underline{0.8102}} & \best{\underline{0.9651}} & \best{\underline{0.9484}} & \best{\underline{0.9500}} & \best{\underline{0.9809}} & \best{\underline{0.5892}} & \best{\underline{0.7248}} & \best{\underline{0.7634}} & \best{\underline{0.6468}} & \best{\underline{0.6773}} & \best{\underline{0.7516}} \\
\midrule
\multicolumn{13}{@{}l}{\textbf{HotpotQA}} \\
SAUP & 0.4573 & 0.6477 & 0.4810 & 0.3902 & 0.4067 & 0.4531 & 0.2532 & 0.4301 & 0.1372 & 0.2546 & 0.2551 & 0.2382 \\
TRACER & 0.4294 & 0.4837 & 0.3613 & 0.3561 & 0.3343 & 0.3281 & 0.2677 & 0.4773 & 0.1486 & 0.2664 & 0.2624 & 0.2626 \\
AUQ & \second{0.5127} & 0.5927 & \second{0.6177} & \second{0.5761} & \second{0.5760} & \second{0.6017} & 0.2808 & \second{0.6013} & \second{0.2235} & \second{0.3264} & \second{0.3337} & \second{0.3337} \\
UProp & 0.4319 & \second{0.6528} & 0.4791 & 0.3927 & 0.4067 & 0.4336 & \second{0.2903} & 0.5811 & 0.1832 & 0.2929 & 0.2996 & 0.3306 \\
\textbf{LProp} & \best{\underline{0.5253}} & \best{\underline{0.7468}} & \best{\underline{0.6638}} & \best{\underline{0.5829}} & \best{\underline{0.6156}} & \best{\underline{0.6602}} & \best{\underline{0.4024}} & \best{\underline{0.8503}} & \best{\underline{0.4539}} & \best{\underline{0.4630}} & \best{\underline{0.4933}} & \best{\underline{0.5637}} \\
\bottomrule
\end{tabularx}
\caption{\textbf{Uncertainty propagation main results.} LProp uses a three-head probability ensemble, with metrics computed after answer selection. All metrics prefer higher values. HotpotQA uses answer-token F1 and exact-match correctness for AUROC, AUARC and retained-answer metrics. The best values use boldface and darker shading; second-best distinct values use lighter shading. Approximate 95\% bootstrap interval half-widths are reported in Appendix~\ref{app:propagation-intervals}, Table~\ref{tab:propagation-intervals}.}
\label{tab:propagation}
\end{table*}

\subsection{LProp}
\label{sec:propagation-experiments}
\subsubsection{Experimental Setting}
\textbf{Benchmarks.}
We evaluate LProp using Qwen3-8B and Mistral-7B-Instruct-v0.3~\citep{mistralai2024mistralv03} on BoolQ~\citep{clark-etal-2019-boolq}, StrategyQA, MMLU mathematics~\citep{hendrycks2021measuringmassivemultitasklanguage}, and HotpotQA, covering yes/no reasoning, mathematics, and open-ended multi-hop QA. We convert each QA instance into a controlled source--receiver episode. The source forms candidate support from private evidence; the receiver obtains the handoff and successive public observations without access to the raw private context. Data splits and protocol details appear in Appendices~\ref{sec:propagation-data-splits} and~\ref{sec:multistep-agent-protocol}. A recorded recovery episode illustrating this setting and LProp's updates appears in Appendix~\ref{app:case-lprop-recovery}.

\noindent\textbf{Compared methods.}
We compare with four representative uncertainty propagation methods: SAUP~\citep{zhao2025uncertainty}, TRACER~\citep{tayebati2026tracer}, AUQ~\citep{zhang2026agenticuq}, and UProp~\citep{duan2025uprop}. Detailed protocols and baseline configurations appear in Appendix~\ref{sec:propagation-protocol}.

\noindent\textbf{Metrics.}
We report F1, AUROC, AUARC, @20, @30, and @50, with higher values indicating better performance. F1 measures token overlap with reference answers. AUROC measures error discrimination, while AUARC summarizes retained-answer accuracy across coverage levels. @20, @30, and @50 measure accuracy after rejecting the highest-risk 20\%, 30\%, and 50\% of answers, respectively.

\subsubsection{Main Results}
\textbf{Propagation performance.}
Table~\ref{tab:propagation} shows that LProp achieves the highest AUROC in all eight evaluated settings, indicating stronger discrimination between correct and erroneous answers after evidence handoffs. This consistent advantage spans the evaluated task types and backbones. LProp also achieves the highest AUARC and all reported retained-correctness scores in every setting. The joint gains in AUARC and individual rejection-rate metrics show that its selective-reliability advantage extends beyond a single operating point. Moreover, LProp can recover from initially incorrect judgments by revising inherited beliefs as new evidence arrives, achieving the highest F1 in all eight settings. Appendix~\ref{app:illustrative-cases} provides a case study illustrating this recovery. These results support the effectiveness of the complete propagation system; Appendix~\ref{app:propagation-results} details gains by setting.

\subsubsection{Cost Analysis}
\label{sec:cost-analysis}
\begin{figure}[!ht]
\centering
\includegraphics[width=\columnwidth]{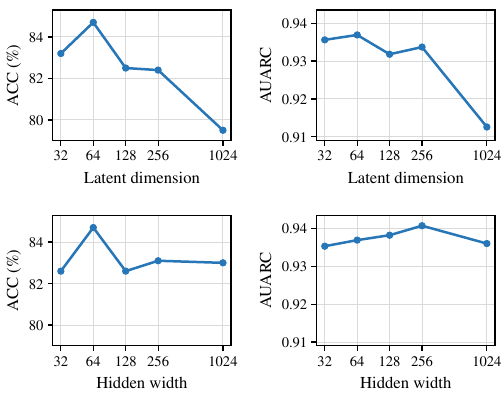}
\caption{\textbf{Sensitivity to model capacity on BoolQ with Qwen3-8B.} Top: latent dimension; bottom: hidden width, with the other parameter fixed at 64. Horizontal axes use a log scale.}
\label{fig:boolq-capacity}
\label{fig:boolq-dimension}
\label{fig:boolq-width}
\end{figure}

\begin{figure}[!ht]
\centering
\includegraphics[width=\columnwidth]{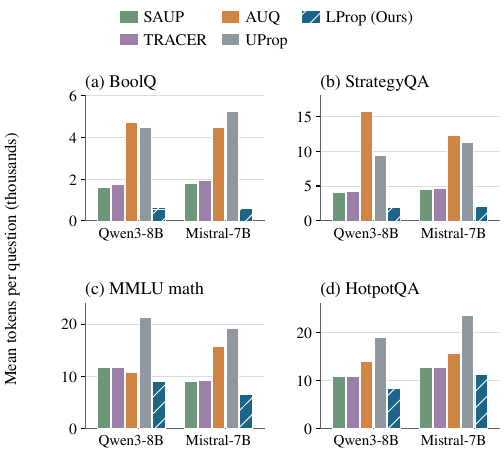}
\caption{\textbf{Inference token cost in the LProp propagation experiment.} Values report mean tokens per question for LProp and the compared baselines. The reported token costs are averaged over three runs.}
\label{fig:token-cost}
\end{figure}

Figure~\ref{fig:token-cost} compares mean inference tokens per question, including generation inputs and outputs, backbone encoding, and auxiliary scoring. LProp uses the fewest tokens in all eight settings, with unweighted mean per-setting reductions of 61.2\% relative to AUQ and 71.1\% relative to UProp. This efficiency stems from reusing candidate representations and applying learned updates to new evidence, avoiding additional reflection or trajectory sampling at each handoff. These savings accompany improved uncertainty assessment across all eight settings. Appendix~\ref{sec:cost-details} provides the accounting protocol and detailed comparisons.

\begin{figure}[!ht]
\centering
\includegraphics[width=\columnwidth]{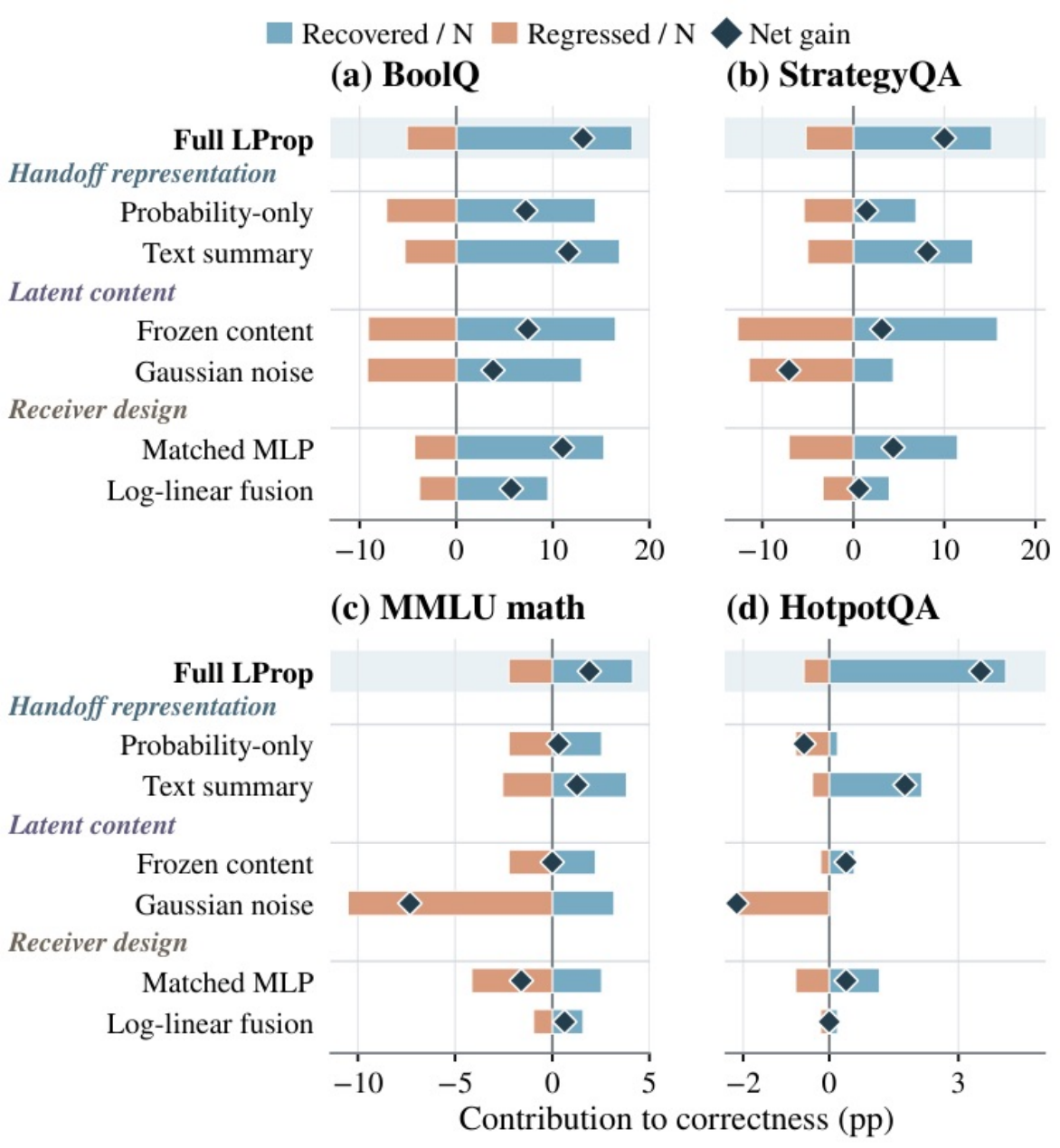}
\caption{\textbf{Correctness gains and losses in matched receiver ablations.} Bars show recovered errors (right) and introduced errors (left), each as a percentage of all questions; diamonds show net gains. Axes vary by benchmark.}
\label{fig:receiver-ablation}
\end{figure}

\subsubsection{Ablation Study}
\label{sec:dimension-ablation}
\label{sec:width-ablation}
We examine the contribution of inherited information and receiver updates, sensitivity to model capacity, and reliability across rejection rates. Details appear in Appendices~\ref{app:receiver-ablation},~\ref{sec:boolq-parameters}, and~\ref{app:rejection-sweep}.

\noindent\textbf{Feature ablations.}
We freeze Qwen3-8B and group the controls by \emph{handoff representation} (probability-only and 128-token text summaries), \emph{latent content} (frozen content and Gaussian noise), and \emph{receiver design} (matched MLP and log-linear fusion). Figure~\ref{fig:receiver-ablation} contrasts recovery of initially wrong answers with regression of initially correct answers. Full LProp achieves the highest final correctness on all four benchmarks and recovers more errors than both alternative handoff representations. On StrategyQA, freezing content recovers slightly more errors but introduces substantially more regressions, highlighting the importance of selective revision. Appendix~\ref{app:receiver-ablation} gives the grouped protocol and counts.

\noindent\textbf{Sensitivity to model capacity.}
Figure~\ref{fig:boolq-capacity} shows that on BoolQ with Qwen3-8B, latent dimension 64 achieves the best ACC and AUARC, while hidden widths 64 and 256 achieve the best ACC and AUARC, respectively. Detailed ablations and configuration selection appear in Appendix~\ref{sec:boolq-parameters}.

\begin{figure}[!ht]
\centering
\includegraphics[width=\columnwidth]{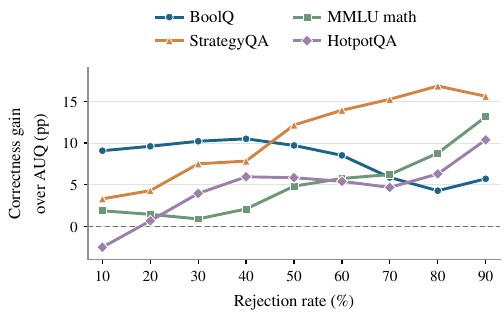}
\caption{\textbf{Rejection-rate sensitivity on Qwen3-8B.} Lines show LProp minus AUQ in retained-answer correctness. Positive values favor LProp; the dashed line denotes equal performance.}
\label{fig:rejection-qwen}
\end{figure}

\noindent\textbf{Rejection-rate sensitivity.}
Figure~\ref{fig:rejection-qwen} compares LProp with AUQ across rejection rates on Qwen3-8B. LProp achieves higher retained-answer correctness at 35 of 36 operating points, supporting its advantage across rejection budgets. Appendix~\ref{app:rejection-sweep} provides the protocol and complete values.

\section{Conclusion}
\label{sec:conclusion}
We introduced \textit{Latent Uncertainty} and validated its ability to predict answer errors, providing an empirical foundation for uncertainty propagation through latent representations. Building on this foundation, \textit{LProp} preserves candidate-specific latent information and performs efficient, evidence-conditioned belief updates, enabling substantial recovery from inherited errors without additional trajectory sampling or textual reflection at each handoff. Matched Feature ablations further demonstrate that inherited latent content improves downstream error recovery beyond answer probabilities or model capacity alone. Across diverse tasks, LProp achieves the highest performance and the lowest mean inference-token usage among the compared methods in all settings.

\newpage
\section*{Limitations}

Our approach requires access to backbone hidden states and the answer scores used to initialize candidate support. Freezing the backbone removes the need to update its parameters, but does not remove this access requirement. Models exposed only through text-generation or token-probability interfaces cannot directly supply the latent features required by the reader and carrier; compatibility therefore depends on internal-state access rather than model size alone.

The latent interface is also tied to the backbone's representation space. Source and evidence projections, feature normalization, and learned heads are fitted using training data for the selected setting. Their parameters and serialized carriers are not inherently interchangeable across backbones, even when the models share an answer format. Changing the backbone or its hidden-state representation may therefore require refitting the associated components.

Our efficiency analysis reports inference-token usage under the specified accounting protocol rather than hardware-independent speedups. Public evidence still requires backbone encoding, while hidden-state extraction, carrier storage, and learned updates introduce computational and storage requirements that token counts do not fully characterize. Actual latency and device-memory usage depend on hardware, model size, batching, and serving implementations. Token savings should therefore not be interpreted as proportional reductions in runtime or memory consumption.

\section*{Ethical Considerations}
This work aims to improve reliability assessment and evidence-based revision in LLM agents. It uses existing benchmarks and pretrained language models, whose reuse remains subject to applicable access and redistribution terms. Consequential applications should include appropriate validation and human oversight. Applications involving sensitive data should include safeguards for access, retention, and reuse. AI was used for writing refinement and code development under human guidance and supervision.

\newpage

\bibliography{references}

\clearpage

\DeclareRobustCommand{\AppendixRQBadge}[1]{%
  \tikz[baseline=(rqtag.base)]{%
    \node[draw=orange!75!black, fill=orange!7,
      rounded corners=2pt, inner xsep=3pt, inner ysep=1pt,
      text=orange!75!black, font=\sffamily\bfseries\small]
      (rqtag) {#1};}}

\providecommand{\AppendixBest}[1]{%
  {\setlength{\fboxsep}{1.2pt}\colorbox{black!16}{\textbf{#1}}}}
\providecommand{\AppendixSecond}[1]{%
  {\setlength{\fboxsep}{1.2pt}\colorbox{black!7}{#1}}}

\appendix

\setcounter{topnumber}{3}
\setcounter{bottomnumber}{2}
\setcounter{totalnumber}{5}
\setcounter{dbltopnumber}{2}
\renewcommand{\topfraction}{0.9}
\renewcommand{\bottomfraction}{0.8}
\renewcommand{\textfraction}{0.08}
\renewcommand{\floatpagefraction}{0.7}
\renewcommand{\dbltopfraction}{0.9}
\renewcommand{\dblfloatpagefraction}{0.7}
\setlength{\textfloatsep}{12pt plus 2pt minus 2pt}
\setlength{\dbltextfloatsep}{12pt plus 2pt minus 2pt}
\setlength{\floatsep}{10pt plus 2pt minus 2pt}
\setlength{\dblfloatsep}{10pt plus 2pt minus 2pt}
\setlength{\intextsep}{10pt plus 2pt minus 2pt}
\makeatletter
\renewcommand{\paragraph}{\@startsection{paragraph}{4}{\z@}{1.0ex plus 0.3ex minus .2ex}{-1em}{\normalsize\bfseries}}
\makeatother


\section{Implementation Details}
\label{sec:method-details}

Our approach is motivated by the hypothesis that latent representations retain uncertainty-relevant information beyond explicit model outputs. We first validate this premise by demonstrating that latent-state-informed uncertainty estimation can reliably predict errors in fixed answers. Building on this evidence, we develop LProp to preserve candidate-specific latent information across agent handoffs and revise inherited beliefs as new evidence arrives. Matched receiver ablations further establish that this latent information contributes to downstream error recovery beyond answer probabilities or model capacity alone. This section details the implementation; training and evaluation are described in Appendix~\ref{sec:training-evaluation}.

\subsection{Latent Uncertainty Estimation}
\label{app:risk-reader}

\paragraph{Feature construction.}
At each trajectory step, the reader combines a 64-dimensional PCA projection of the backbone hidden state with mean and maximum token negative log-likelihood, normalized output length, and an action-parsing indicator, yielding 68 features. For HotpotQA, two additional token statistics increase the dimension to 70.

\paragraph{Risk prediction.}
The conditional filter in Equation~\ref{eq:risk-filter} uses a learned prior and a two-layer network with hidden width 32 and $\tanh$ activation. An additive $2I$ bias in the transition logits encourages state persistence. We use four latent risk components with fixed anchors evenly spaced between $0.001$ and $0.999$. Their posterior probabilities produce a continuous failure-risk estimate through Equation~\ref{eq:risk-readout}.

\subsection{Carrier Construction and Handoff}
\label{app:carrier-details}

\paragraph{Candidate support.}
The carrier in Equation~\ref{eq:latent-carrier} retains admissible answers for BoolQ, StrategyQA, and MMLU mathematics. For HotpotQA, four source samples construct the candidate set; duplicate normalized answers are merged, and an \textsc{other} category represents answers outside the retained candidates. Candidate construction uses neither future evidence nor gold answers. Initial probability masses are derived from source answer logits using temperature scaling, with an optional learned source correction.

\paragraph{Latent representation.}
For binary and multiple-choice tasks, candidate-specific representations are extracted by appending each admissible answer token to the source prompt and reading its final hidden state. Source and public-evidence representations use separate PCA projections. Projected features are standardized and clipped to $[-8,8]$, with 64 dimensions retained per candidate.

The carrier stores candidate identities, probability masses, and candidate-specific latent content, requiring $O(Kd)$ numerical values for $K$ candidates of dimension $d$. At each handoff, the receiver obtains the carrier, question, and subsequent public evidence without accessing the source's raw private context. Unlike scalar or probability-only handoffs, this representation preserves information about how the source supports competing hypotheses.

\paragraph{Answer support and decoding.}
Candidate identities remain fixed during propagation. On HotpotQA, \textsc{other} participates in probability normalization but is excluded from concrete-answer selection. The selected answer is the admissible candidate with the highest probability, and its uncertainty is one minus that probability in the full distribution, without renormalizing after excluding \textsc{other}. Recovery is therefore restricted to correct answers already present in the source support.

\subsection{Evidence-Conditioned Propagation}
\label{app:propagation-update}
\label{app:binary-update}
\label{app:pooling-details}

LProp applies a common evidence-conditioned update to inherited candidate support. As formulated in Equation~\ref{eq:public-pooling}, the update combines an inherited distribution $b_t$ with public-evidence support $p_t^{\mathrm{pub}}$ and applies candidate-specific residuals:
\begin{equation}
\begin{aligned}
\widetilde w_i^t
&=(1-\alpha_t)b_t(i)+\alpha_t p_t^{\mathrm{pub}}(i),\\
w_i^t
&\propto \widetilde w_i^t\exp(\rho_i^t).
\end{aligned}
\end{equation}
Here $\alpha_t$ controls reliance on public evidence, while $\rho_i^t$ adjusts individual hypotheses using inherited latent content and current evidence. The resulting distribution is normalized over the complete retained support.

\paragraph{Task-specific adaptation.}
The same update framework accommodates different answer formats and evidence schedules. For binary QA, candidate support is revised through learned evidence-dependent residuals, with $\alpha_t=0$. The residuals can incorporate changes in public answer log odds, while latent content is updated through the bounded transition in Equation~\ref{eq:joint-update}. Evidence differences are evaluated relative to either the question-only state or the preceding evidence state, as defined in Equation~\ref{eq:evidence-difference}.

For multi-candidate tasks, a learned gate determines $\alpha_t$ from candidate content, inherited probabilities, distribution entropy, and evidence-change features. Candidate-specific residuals further adjust the resulting support. Depending on the task, updates either carry forward revised content and probabilities or retain the source state while incorporating cumulative public evidence. These adaptations share the same carrier, candidate support, and probability-update interface; they differ only in how evidence is incorporated into the common revision mechanism.


\section{Training and Evaluation Details}
\label{sec:training-evaluation}

\subsection{Latent Uncertainty Estimation}
\label{sec:estimator-protocol}

\subsubsection{Evaluation Protocol}

\paragraph{Agent execution.}
Each question is evaluated using a fixed greedy ReAct trajectory containing tool observations and the final answer, with at most six interaction steps. All estimators assess the same fixed answers, while auxiliary sampled trajectories provide additional uncertainty observations without altering the evaluated answers.

\paragraph{Data splits.}
Table~\ref{tab:estimator-data-splits} reports the training, development, and evaluation sizes for each setting. Training data are used to fit the reader, while development data support model selection and calibration.

\begin{table}[!htbp]
\centering
\small
\setlength{\tabcolsep}{3pt}
\begin{tabular}{llrrr}
\toprule
Dataset & Backbone & Train & Dev. & Eval. \\
\midrule
HotpotQA & Qwen3-8B & 512 & 128 & 1,024 \\
HotpotQA & Qwen2.5-7B & 512 & 128 & 1,024 \\
StrategyQA & Qwen3-8B & 512 & 128 & 1,024 \\
\bottomrule
\end{tabular}
\caption{Question counts for the Latent Uncertainty Estimator. Each row represents one dataset--backbone setting.}
\label{tab:estimator-data-splits}
\end{table}

\subsubsection{Reader Training}
\label{app:reader-training}

The reader is trained with binary cross-entropy against answer-error labels while keeping evaluated answers fixed. This supervision directly assesses whether latent-state-informed features can predict answer failures independently of answer format. PCA and feature normalization are fitted on training data and frozen during evaluation. Backbone weights remain frozen. The reader is trained with five seeds: 17, 29, 43, 71, and 101. Each trained reader is evaluated independently, and the reported metrics are averaged across the five runs.

\subsubsection{Baseline Configurations}

\paragraph{Token Entropy.}
Token Entropy averages the entropy of the next-token probability distribution across generation positions, measuring how diffuse the model's token predictions are~\citep{fomicheva2020unsupervised}.

\paragraph{Mean token NLL.}
Mean token NLL averages the negative log-probabilities assigned to the generated tokens, providing a length-normalized measure of how unlikely the observed generation is~\citep{fomicheva2020unsupervised}.

\paragraph{Sequence NLL.}
Sequence NLL sums the negative log-probabilities of the generated tokens to score the entire sequence without length normalization~\citep{malinin2020uncertainty}.

\paragraph{Predictive Entropy.}
Predictive Entropy estimates entropy over output sequences by averaging the negative log-probabilities of sampled generations~\citep{malinin2020uncertainty,kuhn2023semantic}. Unlike Sequence NLL, it assesses uncertainty across sampled outputs rather than scoring a single generation.

\paragraph{Length-normalized Predictive Entropy.}
This variant divides each sampled sequence's negative log-probability by its token length before averaging, reducing the influence of generation length on the uncertainty estimate~\citep{malinin2020uncertainty}.

\paragraph{Semantic Entropy.}
Semantic Entropy groups sampled answers by semantic equivalence and estimates entropy over the resulting meaning-level distribution~\citep{kuhn2023semantic}. It distinguishes disagreement in answer meaning from variation in wording.

The three sampling-based baselines each use ten sampled trajectories.

\paragraph{Semantic Entropy Probe.}
Semantic Entropy Probe~\citep{kossen2024semanticentropyprobesrobust} predicts the Semantic Entropy score and uses the predicted value as its uncertainty score. It is denoted as Semantic Probe in the figure and table. The prediction target is semantic entropy, whereas our latent reader is trained on answer-error labels. AUROC, AUARC, and @30 use the probe score to rank uncertainty; Brier uses the failure probability obtained from the development-fitted sigmoid calibration in Appendix~\ref{sec:estimator-calibration}. The remaining evaluation settings, including the fixed evaluated answers and calibration protocol, are the same as for Semantic Entropy.

\subsubsection{Probability Calibration}
\label{sec:estimator-calibration}

Figure~\ref{fig:estimation} and Table~\ref{tab:estimation} report development-calibrated Brier scores. All methods use the same two-parameter sigmoid calibration with a positive slope. Inputs are standardized using development data, with native risk scores for the baselines and failure-probability logits for the latent reader.

Calibration uses 118 development examples for HotpotQA/Qwen3, 103 for HotpotQA/Qwen2.5, and 123 for StrategyQA/Qwen3. The calibration parameters are fitted on development data and frozen before evaluation. Each reader seed is calibrated independently, and its Brier score is included in the five-run average.

The calibration objective is mean binary log loss with a slope penalty of $0.5\times10^{-3}a^2$, where $a\geq10^{-6}$ is the sigmoid slope and the intercept is unpenalized. Both parameters are optimized using L-BFGS-B. The positive slope preserves the original risk rankings and ties, leaving AUROC, AUARC, and @30 unchanged while enabling Brier evaluation on a common probability scale.

\subsubsection{Evaluation Metrics and Results}
\label{sec:complete-estimation}
\label{app:estimator-results}

Table~\ref{tab:estimation} contains all values behind Figure~\ref{fig:estimation}, using the 1,024-question evaluation cohorts in Table~\ref{tab:estimator-data-splits}. Calibration follows Appendix~\ref{sec:estimator-calibration}.

\paragraph{Reading the fixed-answer comparison.}
The estimators share the same evaluated answers, which holds answer correctness fixed when comparing uncertainty estimates. AUROC measures the separation of erroneous and correct answers by estimated risk; AUARC summarizes retained-answer accuracy across coverage levels. The two metrics provide complementary views of ranking quality: AUROC compares error--correct answer pairs, while AUARC evaluates the subsets selected at successive coverage levels. A change in the ordering can consequently affect AUROC and AUARC differently, because they summarize different aspects of the same selection problem. The @30 result identifies a particular operating point, retaining the lowest-risk 70\% of answers. Calibrated Brier scores evaluate the quality of the failure probabilities obtained through development-set calibration. Together, these metrics characterize error discrimination, selective correctness, and probability prediction.

\begin{table*}[!tp]
\centering
\setlength{\tabcolsep}{3pt}
\fontsize{8}{9.5}\selectfont
\resizebox{\textwidth}{!}{%
\begin{tabular}{lrrrrrrrrrrrr}
\toprule
& \multicolumn{4}{c}{HotpotQA / Qwen3-8B}
& \multicolumn{4}{c}{HotpotQA / Qwen2.5-7B}
& \multicolumn{4}{c}{StrategyQA / Qwen3-8B} \\
\cmidrule(lr){2-5}\cmidrule(lr){6-9}\cmidrule(lr){10-13}
Method
& AUROC $\uparrow$ & AUARC $\uparrow$ & @30 $\uparrow$ & Brier $\downarrow$
& AUROC $\uparrow$ & AUARC $\uparrow$ & @30 $\uparrow$ & Brier $\downarrow$
& AUROC $\uparrow$ & AUARC $\uparrow$ & @30 $\uparrow$ & Brier $\downarrow$ \\
\midrule
Semantic Ent.
& .7534 & .7994 & 80.56 & .1677
& .8296 & .7335 & 70.43 & .1579
& .6923 & .7528 & 74.48 & .1946 \\
Semantic Probe
& .6903 & .7781 & 75.12 & .1824
& .7884 & .7012 & 68.53 & .1722
& .6703 & .7392 & 72.18 & .2136 \\
Token Ent.
& .7128 & .7937 & 76.15 & .1812
& .7984 & .7118 & 68.76 & .1703
& .6599 & .7456 & 72.25 & .2018 \\
Predictive Ent.
& .7005 & .7758 & 76.43 & .1866
& .7901 & .7054 & 67.92 & .1718
& .6935 & .7673 & 73.92 & .1972 \\
Len.-norm. PE
& .7199 & .7902 & 76.43 & .1822
& .7943 & .7077 & 68.06 & .1689
& .6789 & .7584 & 72.38 & .2008 \\
Mean token NLL
& .7039 & .7890 & 76.15 & .1830
& .7912 & .7093 & 68.48 & .1714
& .6532 & .7420 & 71.41 & .2030 \\
Sequence NLL
& .6890 & .7760 & 75.31 & .1871
& .7963 & .7128 & 68.90 & .1704
& .6664 & .7523 & 72.25 & .2005 \\
\textbf{Ours}
& \textbf{.8079} & \textbf{.8317} & \textbf{83.38} & \textbf{.1533}
& \textbf{.9032} & \textbf{.7911} & \textbf{73.44} & \textbf{.1172}
& \textbf{.7189} & \textbf{.7837} & \textbf{74.67} & \textbf{.1942} \\
\bottomrule
\end{tabular}}
\caption{\textbf{Complete uncertainty estimation results.} Bold values indicate the best result within each setting and metric. @30 is a percentage; other metrics use their original scales. Brier uses development-calibrated failure probabilities.}
\label{tab:estimation}
\end{table*}

Figure~\ref{fig:estimation} shows that the Latent Uncertainty Estimator achieves the best point estimates across all four metrics in all three settings. On HotpotQA, AUROC reaches 0.8079 with Qwen3 and 0.9032 with Qwen2.5, compared with 0.7534 and 0.8296 for Semantic Entropy. On StrategyQA, it reaches 0.7189, compared with 0.6935 for Predictive Entropy.

The improved rankings also retain more correct answers at a fixed rejection budget. HotpotQA @30 increases from 80.56\% to 83.38\% on Qwen3 and from 70.43\% to 73.44\% on Qwen2.5 relative to Semantic Entropy. AUARC improves from 0.7994 to 0.8317 and from 0.7335 to 0.7911, respectively, showing that the benefit extends across coverage levels. On StrategyQA, AUARC rises from the strongest baseline's 0.7673 to 0.7837, while @30 improves from 74.48\% to 74.67\%.

Calibrated Brier scores complement these ranking results. On HotpotQA, Brier decreases from 0.1677 to 0.1533 with Qwen3 and from 0.1579 to 0.1172 with Qwen2.5. Table~\ref{tab:estimation} reports the complete Brier results alongside the discrimination and retention metrics.

\subsection{Uncertainty Propagation}
\label{sec:propagation-protocol}

\subsubsection{Data Splits and Evaluation Cohorts}
\label{sec:propagation-data-splits}

Table~\ref{tab:propagation-data-splits} reports the training, development, and evaluation question counts for each dataset--backbone setting. All compared methods use the same evaluation questions within each setting.

\begin{table}[!htbp]
\centering
\small
\setlength{\tabcolsep}{3pt}
\begin{tabular}{llrrr}
\toprule
Dataset & Backbone & Train & Dev. & Eval. \\
\midrule
BoolQ & Qwen3-8B & 4,096 & 512 & 1,000 \\
 & Mistral-7B & 4,096 & 512 & 1,000 \\
StrategyQA & Qwen3-8B & 512 & 128 & 480 \\
 & Mistral-7B & 512 & 128 & 480 \\
MMLU math & Qwen3-8B & 512 & 128 & 314 \\
 & Mistral-7B & 512 & 128 & 314 \\
HotpotQA & Qwen3-8B & 512 & 128 & 512 \\
 & Mistral-7B & 512 & 128 & 512 \\
\bottomrule
\end{tabular}
\caption{Question counts for the LProp main experiments. Mistral-7B denotes Mistral-7B-Instruct-v0.3. For both backbones on MMLU math, training and development use MMLU-Pro math.}
\label{tab:propagation-data-splits}
\end{table}

\paragraph{Training and development.}
Both backbones use the same split settings within each dataset. BoolQ training and development questions come from the official training split. Questions or passages overlapping official validation are excluded, and duplicates are grouped before deterministic hash-based splitting. StrategyQA uses 512 training and 128 development questions. The mathematics heads are trained and selected on MMLU-Pro math~\citep{wang2024mmlu} before evaluation on MMLU mathematics. HotpotQA training and development sets are disjoint, question-type-stratified subsets of the official training split, with normalized duplicate questions and overlap with official validation excluded. Model fitting and feature transformations use training data, while model selection uses development data.

\paragraph{Evaluation cohorts.}
BoolQ uses 1,000 questions from official validation per backbone. StrategyQA uses the 480 questions with subsequent public evidence from a collected set of 512. Both backbones share the same 314-question MMLU mathematics cohort, comprising 191 official test questions and 123 official development/validation questions across abstract algebra, college mathematics, elementary mathematics, high-school mathematics, and high-school statistics. This cohort is deduplicated against the previously used MMLU-Pro math training, development, and evaluation questions. The 123 official development/validation questions are used for evaluation, not head selection, which uses the separate MMLU-Pro development set. HotpotQA uses 512 question-type-stratified questions from official distractor validation, shared across backbones.

\subsubsection{QA-to-Multi-Step Agent Construction}
\label{sec:multistep-agent-protocol}

\paragraph{Episode structure and information boundary.}
Each QA instance is organized into an ordered source--receiver episode while retaining its original question and answer criterion. The source processes the question and private evidence or a reasoning draft to construct candidate support. It then transmits a carrier containing candidate identities, probability masses, and candidate-specific latent content. The receiver processes subsequent public evidence without accessing the source's raw private text, future evidence, or reference answer. Source and receiver are roles separated by this information boundary within the same backbone setting.

Each subsequent public evidence event defines an update stage. The frozen backbone encodes the available public context, and LProp uses these evidence features and inherited latent information to revise support over the fixed candidate set.

\paragraph{BoolQ: controlled passage release.}
The original passage is divided into three contiguous blocks at approximately one third and two thirds of its sentence sequence. Passages with fewer than three sentences are divided by words instead. The first block is private to the source, while the second and third blocks are revealed sequentially. The first receiver update uses the question and block 2; the second uses the question and blocks 2 and 3. Block 1 is represented only through the carrier. Splitting preserves the original order and uses neither answer labels nor sentence relevance. Candidate support remains \{\textit{no}, \textit{yes}\} throughout.

\paragraph{StrategyQA: replayed retrieval observations.}
A Qwen3-8B collector generates greedy ReAct trajectories using \texttt{Search} and \texttt{Lookup} over a fixed Wikipedia corpus. Search attempts exact-title matching followed by BM25 retrieval, while Lookup returns the next matching sentence from the first passage in the latest search result. Retrieval prompts exclude gold facts and decompositions.

The first parsed, nonterminal retrieval observation becomes private source evidence, and the source is encoded after receiving it. Later observations become public events in their original order; the collector's final answer is not included as evidence. Both backbones receive the same frozen observation sequence and encode it separately. Public contexts accumulate the observations revealed so far while excluding the private observation. Repeated observations are deduplicated, and questions without new content after deduplication remain in the evaluation cohort.

\paragraph{MMLU mathematics: draft, derivation, and check.}
Each question retains its original answer options and follows three information stages: a private generated draft, a separately generated public derivation, and a public check of that derivation. The source processes the private draft and constructs an option-indexed carrier. The public derivation and check then arrive as successive evidence events, and the receiver revises support over the same options after each arrival. The reasoning segments are generated greedily and reused across compared methods. Derivations and checks are treated as fallible evidence, while reference answers are reserved for supervision and scoring.

\paragraph{HotpotQA: staged passage release.}
A question-conditioned BM25 retriever ranks the distractor passages without using answer or supporting-fact labels. For $n$ passages, the top $\min(5,\lceil n/2\rceil)$ form the source's private context. The remaining passages are divided into two disjoint, nearly equal public groups, with the first group larger when the remainder is odd. These groups are revealed sequentially; empty groups cause no update. A ten-passage question therefore has five private passages followed by public groups of three and two passages.

The source generates four answers from its private context. Normalized duplicates are merged into retained candidates, and an \textsc{other} category represents answers outside this support. At each public arrival, LProp updates candidate support using the available evidence. Private passages are not forwarded to the receiver, and the final answer is selected from the retained concrete candidates.

\paragraph{Execution and readout.}
The carrier is serialized and deserialized before receiver-side processing. We evaluate the final receiver state after all available public evidence has been processed. Answer selection follows Appendix~\ref{app:carrier-details}.

\subsubsection{LProp Training}
\label{app:training-readout}

\paragraph{Supervision.}
LProp is trained on available handoff views using the objective in Equation~\ref{eq:propagation-loss}, which combines negative log-likelihood of the target candidate with a Brier penalty. For binary QA, the Brier term measures squared error on the yes probability; for categorical tasks, it measures squared distance to the one-hot target. HotpotQA uses the matching source candidate or \textsc{other} as its target. Minibatches are organized by question to prevent unequal numbers of evidence views from dominating training. Where enabled, an unrelated-evidence consistency penalty discourages unjustified changes to inherited support.

\paragraph{Optimization and inference.}
The backbone remains frozen, and propagation heads are optimized with AdamW using weight decay $0.01$ and gradient-norm clipping at $1.0$. Training consists of initial fitting followed by lower-learning-rate refinement. Source and public-evidence PCA projections, feature normalization, and the source probability temperature are fitted on training data and frozen before evaluation. Development data are used for model selection. At inference, candidate probabilities from three independently trained heads are averaged before answer selection. Public evidence is encoded by the frozen backbone, and the learned update revises inherited support without additional trajectory sampling or textual reflection at each handoff.

\subsubsection{Baseline Configurations}

We implement each baseline according to its original formulation and released procedures, applying training only where required by the method.

\paragraph{SAUP.}
SAUP~\citep{zhao2025uncertainty} aggregates step-level uncertainty using situation-aware weights that account for the agent's reasoning and its alignment with observations. We use its direct-distance weighting configuration.

\paragraph{TRACER.}
TRACER~\citep{tayebati2026tracer} combines token-level surprisal with repetition and coherence indicators, using tail-focused aggregation to emphasize critical high-risk episodes. We use its QA encoder adaptation.

\paragraph{UProp.}
UProp~\citep{duan2025uprop} decomposes decision uncertainty into intrinsic uncertainty and uncertainty inherited from preceding decisions. Its distribution-level estimator uses sampled trajectories and alternative decisions to estimate the inherited component.

\paragraph{AUQ.}
AUQ~\citep{zhang2026agenticuq} retains verbalized confidence and explanations in memory and uses uncertainty-triggered reflection to revise tentative decisions. We preserve its original memory and reflection procedure.

\paragraph{Shared protocol and uncertainty-guided reflection.}
All methods are evaluated on the same questions and evidence schedule, with required auxiliary inference included in token accounting. SAUP, TRACER, and UProp provide post-hoc uncertainty estimates rather than an answer-revision mechanism. We therefore append one final reflection step to each method: the agent receives its trajectory-level uncertainty estimate and reflects once before producing the final answer used for evaluation.

\subsubsection{Metrics and Answer Handling}
\label{sec:propagation-metrics}

F1 is the mean answer-token F1 across questions. For canonical yes/no and option answers, it equals correctness. On HotpotQA, token F1 awards partial credit for overlap with the reference answer, while AUROC, AUARC, and retained-answer correctness use exact match. Erroneous answers are the positive class for AUROC. AUARC averages retained-answer correctness across coverage levels; @20, @30, and @50 report correctness after rejecting the highest-risk 20\%, 30\%, and 50\% of answers, respectively.

All questions, including those without a matching source candidate on HotpotQA, remain in the evaluation denominator.

\subsubsection{Detailed Result Comparisons}
\label{app:propagation-results}

Table~\ref{tab:propagation} shows that LProp achieves the highest AUROC, AUARC, and all three retained-correctness metrics in every dataset--backbone setting. It also achieves the highest F1 in all eight settings. Relative to the strongest baseline in each setting, AUROC gains range from 0.0816 to 0.2752 and AUARC gains from 0.0388 to 0.4305. These results indicate improvements in both answer quality and selective reliability across the evaluated settings.

On BoolQ, LProp raises F1 from the strongest baseline's 78.70\% to 83.10\% with Qwen3 and from 71.90\% to 80.10\% with Mistral. The corresponding @30 values reach 91.86\% and 88.29\%, exceeding the strongest baselines by 10.22 and 14.14 percentage points. On MMLU mathematics with Mistral, F1 increases from 46.82\% to 58.92\%, while AUARC rises from 0.3329 to 0.7634. With Qwen3, @50 reaches 98.09\%, compared with 93.26\% for the strongest baseline.

On HotpotQA/Mistral, LProp improves F1 over AUQ from 28.08\% to 40.24\% and AUARC from 0.2235 to 0.4539. On HotpotQA/Qwen3, LProp achieves the highest AUROC, AUARC, and reported selective-correctness scores; for example, @30 is 61.56\%, compared with 57.60\% for AUQ. These results extend the selective-reliability gains to open-ended multi-hop QA.

\paragraph{Backbone access and model scale.}
LProp requires white-box access to backbone hidden states and candidate answer scores. We therefore use locally deployed Qwen3-8B and Mistral-7B-Instruct-v0.3, whose sizes fit our available computational resources. This choice reflects deployment constraints rather than an inherent model-size requirement of LProp. The consistent gains across these backbones show that useful latent uncertainty can support evidence-conditioned revision even at this scale. These findings motivate the expectation that larger backbones can also benefit from the same mechanism, provided that their internal states are accessible and the projections and propagation heads are fitted to their representations.


\subsubsection{Bootstrap Intervals}
\label{app:propagation-intervals}
Table~\ref{tab:propagation-intervals} reports half-widths $h=1.96\times\mathrm{SE}_{\mathrm{boot}}$, where $\mathrm{SE}_{\mathrm{boot}}$ is the bootstrap standard error. For the corresponding main-table estimate $\hat{m}$, the approximate 95\% interval is $[\hat{m}-h,\hat{m}+h]$. These symmetric normal-approximation intervals are not constrained to $[0,1]$.

\begin{table*}[!tp]
\centering
\fontsize{7.5}{9.2}\selectfont
\setlength{\tabcolsep}{1.2pt}
\renewcommand{\arraystretch}{1.08}
\begin{tabularx}{\textwidth}{@{}l*{12}{>{\centering\arraybackslash}X}@{}}
\toprule
& \multicolumn{6}{c}{\textbf{Qwen3-8B}}
& \multicolumn{6}{c}{\textbf{Mistral-7B-Instruct-v0.3}} \\
\cmidrule(lr){2-7}\cmidrule(lr){8-13}
Method & F1 & AUROC & AUARC & @20 & @30 & @50
& F1 & AUROC & AUARC & @20 & @30 & @50 \\
\midrule
\multicolumn{13}{@{}l}{\textbf{BoolQ}} \\
SAUP & $\pm 0.0251$ & $\pm 0.0428$ & $\pm 0.0303$ & $\pm 0.0272$ & $\pm 0.0287$ & $\pm 0.0325$ & $\pm 0.0217$ & $\pm 0.0377$ & $\pm 0.0333$ & $\pm 0.0317$ & $\pm 0.0337$ & $\pm 0.0387$ \\
TRACER & $\pm 0.0234$ & $\pm 0.0432$ & $\pm 0.0346$ & $\pm 0.0283$ & $\pm 0.0301$ & $\pm 0.0349$ & $\pm 0.0287$ & $\pm 0.0381$ & $\pm 0.0378$ & $\pm 0.0319$ & $\pm 0.0344$ & $\pm 0.0405$ \\
AUQ & $\pm 0.0265$ & $\pm 0.0318$ & $\pm 0.0214$ & $\pm 0.0272$ & $\pm 0.0272$ & $\pm 0.0254$ & $\pm 0.0293$ & $\pm 0.0294$ & $\pm 0.0299$ & $\pm 0.0304$ & $\pm 0.0306$ & $\pm 0.0306$ \\
UProp & $\pm 0.0250$ & $\pm 0.0462$ & $\pm 0.0363$ & $\pm 0.0266$ & $\pm 0.0287$ & $\pm 0.0341$ & $\pm 0.0235$ & $\pm 0.0376$ & $\pm 0.0344$ & $\pm 0.0316$ & $\pm 0.0321$ & $\pm 0.0368$ \\
\textbf{LProp} & $\boldsymbol{\pm 0.0234}$ & $\boldsymbol{\pm 0.0340}$ & $\boldsymbol{\pm 0.0133}$ & $\boldsymbol{\pm 0.0218}$ & $\boldsymbol{\pm 0.0215}$ & $\boldsymbol{\pm 0.0211}$ & $\boldsymbol{\pm 0.0241}$ & $\boldsymbol{\pm 0.0341}$ & $\boldsymbol{\pm 0.0169}$ & $\boldsymbol{\pm 0.0242}$ & $\boldsymbol{\pm 0.0248}$ & $\boldsymbol{\pm 0.0239}$ \\
\midrule
\multicolumn{13}{@{}l}{\textbf{StrategyQA}} \\
SAUP & $\pm 0.0416$ & $\pm 0.0548$ & $\pm 0.0590$ & $\pm 0.0485$ & $\pm 0.0515$ & $\pm 0.0601$ & $\pm 0.0236$ & $\pm 0.0533$ & $\pm 0.0590$ & $\pm 0.0484$ & $\pm 0.0530$ & $\pm 0.0626$ \\
TRACER & $\pm 0.0311$ & $\pm 0.0543$ & $\pm 0.0585$ & $\pm 0.0474$ & $\pm 0.0516$ & $\pm 0.0571$ & $\pm 0.0419$ & $\pm 0.0526$ & $\pm 0.0579$ & $\pm 0.0493$ & $\pm 0.0520$ & $\pm 0.0618$ \\
AUQ & $\pm 0.0426$ & $\pm 0.0528$ & $\pm 0.0515$ & $\pm 0.0449$ & $\pm 0.0483$ & $\pm 0.0557$ & $\pm 0.0331$ & $\pm 0.0471$ & $\pm 0.0505$ & $\pm 0.0458$ & $\pm 0.0489$ & $\pm 0.0562$ \\
UProp & $\pm 0.0237$ & $\pm 0.0568$ & $\pm 0.0634$ & $\pm 0.0476$ & $\pm 0.0520$ & $\pm 0.0637$ & $\pm 0.0436$ & $\pm 0.0531$ & $\pm 0.0610$ & $\pm 0.0498$ & $\pm 0.0517$ & $\pm 0.0603$ \\
\textbf{LProp} & $\boldsymbol{\pm 0.0397}$ & $\boldsymbol{\pm 0.0507}$ & $\boldsymbol{\pm 0.0391}$ & $\boldsymbol{\pm 0.0432}$ & $\boldsymbol{\pm 0.0442}$ & $\boldsymbol{\pm 0.0483}$ & $\boldsymbol{\pm 0.0426}$ & $\boldsymbol{\pm 0.0506}$ & $\boldsymbol{\pm 0.0497}$ & $\boldsymbol{\pm 0.0464}$ & $\boldsymbol{\pm 0.0496}$ & $\boldsymbol{\pm 0.0573}$ \\
\midrule
\multicolumn{13}{@{}l}{\textbf{MMLU}} \\
SAUP & $\pm 0.0314$ & $\pm 0.0555$ & $\pm 0.0371$ & $\pm 0.0357$ & $\pm 0.0362$ & $\pm 0.0419$ & $\pm 0.0565$ & $\pm 0.0644$ & $\pm 0.0748$ & $\pm 0.0630$ & $\pm 0.0664$ & $\pm 0.0798$ \\
TRACER & $\pm 0.0322$ & $\pm 0.0478$ & $\pm 0.0463$ & $\pm 0.0417$ & $\pm 0.0416$ & $\pm 0.0503$ & $\pm 0.0539$ & $\pm 0.0647$ & $\pm 0.0790$ & $\pm 0.0632$ & $\pm 0.0675$ & $\pm 0.0796$ \\
AUQ & $\pm 0.0350$ & $\pm 0.0403$ & $\pm 0.0261$ & $\pm 0.0264$ & $\pm 0.0261$ & $\pm 0.0274$ & $\pm 0.0548$ & $\pm 0.0426$ & $\pm 0.0612$ & $\pm 0.0632$ & $\pm 0.0652$ & $\pm 0.0639$ \\
UProp & $\pm 0.0354$ & $\pm 0.0636$ & $\pm 0.0411$ & $\pm 0.0339$ & $\pm 0.0365$ & $\pm 0.0417$ & $\pm 0.0511$ & $\pm 0.0524$ & $\pm 0.0722$ & $\pm 0.0647$ & $\pm 0.0670$ & $\pm 0.0830$ \\
\textbf{LProp} & $\boldsymbol{\pm 0.0346}$ & $\boldsymbol{\pm 0.0317}$ & $\boldsymbol{\pm 0.0209}$ & $\boldsymbol{\pm 0.0279}$ & $\boldsymbol{\pm 0.0293}$ & $\boldsymbol{\pm 0.0283}$ & $\boldsymbol{\pm 0.0550}$ & $\boldsymbol{\pm 0.0550}$ & $\boldsymbol{\pm 0.0560}$ & $\boldsymbol{\pm 0.0609}$ & $\boldsymbol{\pm 0.0632}$ & $\boldsymbol{\pm 0.0751}$ \\
\midrule
\multicolumn{13}{@{}l}{\textbf{HotpotQA}} \\
SAUP & $\pm 0.0310$ & $\pm 0.0494$ & $\pm 0.0582$ & $\pm 0.0449$ & $\pm 0.0494$ & $\pm 0.0583$ & $\pm 0.0217$ & $\pm 0.0215$ & $\pm 0.0251$ & $\pm 0.0209$ & $\pm 0.0209$ & $\pm 0.0259$ \\
TRACER & $\pm 0.0315$ & $\pm 0.0531$ & $\pm 0.0592$ & $\pm 0.0454$ & $\pm 0.0483$ & $\pm 0.0556$ & $\pm 0.0275$ & $\pm 0.0267$ & $\pm 0.0297$ & $\pm 0.0232$ & $\pm 0.0250$ & $\pm 0.0303$ \\
AUQ & $\pm 0.0360$ & $\pm 0.0443$ & $\pm 0.0526$ & $\pm 0.0376$ & $\pm 0.0394$ & $\pm 0.0391$ & $\pm 0.0295$ & $\pm 0.0334$ & $\pm 0.0342$ & $\pm 0.0346$ & $\pm 0.0347$ & $\pm 0.0347$ \\
UProp & $\pm 0.0396$ & $\pm 0.0491$ & $\pm 0.0594$ & $\pm 0.0459$ & $\pm 0.0486$ & $\pm 0.0582$ & $\pm 0.0238$ & $\pm 0.0303$ & $\pm 0.0272$ & $\pm 0.0251$ & $\pm 0.0279$ & $\pm 0.0336$ \\
\textbf{LProp} & $\boldsymbol{\pm 0.0378}$ & $\boldsymbol{\pm 0.0406}$ & $\boldsymbol{\pm 0.0513}$ & $\boldsymbol{\pm 0.0423}$ & $\boldsymbol{\pm 0.0431}$ & $\boldsymbol{\pm 0.0450}$ & $\boldsymbol{\pm 0.0345}$ & $\boldsymbol{\pm 0.0352}$ & $\boldsymbol{\pm 0.0612}$ & $\boldsymbol{\pm 0.0425}$ & $\boldsymbol{\pm 0.0450}$ & $\boldsymbol{\pm 0.0563}$ \\
\bottomrule
\end{tabularx}
\caption{\textbf{Approximate 95\% interval half-widths for Table~\ref{tab:propagation}.} All half-widths use the same 0--1 scale as the corresponding main-table estimates.}
\label{tab:propagation-intervals}
\end{table*}

\section{Ablation Studies}
\label{sec:ablation-studies}

\subsection{BoolQ Parameter Ablations}
\label{sec:boolq-parameters}

\subsubsection{Sweep Settings}

We evaluate the sensitivity of LProp to latent dimension and hidden width on BoolQ using Qwen3-8B, with 4,096 training, 512 development, and 1,000 evaluation questions. Both sweeps consider values of 32, 64, 128, 256, and 1,024. The latent-dimension sweep fixes hidden width at 64, while the hidden-width sweep fixes latent dimension at 64.

Training configurations remain fixed across each sweep, with projections and normalization fitted on training data only. Predictions are ensembled over three training seeds (17, 29, and 43). We report accuracy (ACC) and AUARC, where ACC equals answer F1 on BoolQ.

\subsubsection{Detailed Results}

Table~\ref{tab:boolq-parameter-ablation} presents the complete results.

\begin{table}[!htbp]
\centering
\small
\setlength{\tabcolsep}{3pt}
\begin{tabular}{@{}rcccc@{}}
\toprule
& \multicolumn{2}{c}{Latent dimension}
& \multicolumn{2}{c}{Hidden width} \\
\cmidrule(lr){2-3}\cmidrule(lr){4-5}
Value & ACC (\%) & AUARC & ACC (\%) & AUARC \\
\midrule
32   & 83.20 & 0.9356 & 82.60 & 0.9353 \\
64   & \textbf{84.70} & \textbf{0.9369}
     & \textbf{84.70} & 0.9369 \\
128  & 82.50 & 0.9318 & 82.60 & 0.9382 \\
256  & 82.40 & 0.9337 & 83.10 & \textbf{0.9407} \\
1024 & 79.50 & 0.9126 & 83.00 & 0.9360 \\
\bottomrule
\end{tabular}
\caption{\textbf{BoolQ parameter ablations on Qwen3-8B.} The latent-dimension sweep fixes hidden width at 64, while the hidden-width sweep fixes latent dimension at 64. Higher values are better for both metrics. Bold indicates the best result within each sweep and metric.}
\label{tab:boolq-parameter-ablation}
\label{tab:boolq-parameters}
\end{table}

\paragraph{Latent dimension.}
A dimension of 64 achieves the highest ACC (84.70\%) and AUARC (0.9369). Compared with dimension 1,024, it improves ACC by 5.20 percentage points and AUARC by 0.0243, indicating that compact latent representations perform better under the evaluated configuration.

\paragraph{Hidden width.}
Width 64 achieves the highest ACC (84.70\%), while width 256 achieves the highest AUARC (0.9407), with ACC of 83.10\%. These results show that full-coverage accuracy and selective reliability favor different network capacities.

\subsubsection{Model Selection}

Configurations are selected using development-set ranks across AUROC, ACC, AUARC, @20, @30, and @50. The selected configuration uses latent dimension 64 and hidden width 256. This configuration is also used in the main BoolQ propagation experiment, evaluated on the same 1,000-question cohort. Its ACC of 83.10\% and AUARC of 0.9407 therefore appear in both the parameter ablation and main results.

\subsection{Sensitivity to the Rejection Rate}
\label{app:rejection-sweep}

We compare LProp and AUQ on Qwen3-8B across BoolQ, StrategyQA, MMLU mathematics, and HotpotQA. The sweep covers rejection rates from 10\% to 90\% in increments of 10 percentage points, yielding 36 paired operating points. We reuse the main experiment's frozen predictions and risk scores without retraining, regenerating answers, or recalibrating scores at individual rates.

At @k, each method retains the $\lceil N(1-k/100)\rceil$ lowest-risk answers, using expected correctness to resolve boundary ties. Each method is evaluated on its own final answers and risk scores. Correctness follows the main protocol; HotpotQA uses exact match rather than answer-token F1. Table~\ref{tab:rejection-qwen} reports all values, including @20, @30, and @50 from the main table. Rates within a benchmark reuse the same predictions and are not independent replications.

Figure~\ref{fig:rejection-qwen} plots LProp minus AUQ in percentage points. LProp has higher retained-answer correctness at 35 of 36 operating points: all nine rates on BoolQ, StrategyQA, and MMLU mathematics, and @20--@90 on HotpotQA. These results extend the comparison beyond the three rates in the main table. Answer handling follows Appendix~\ref{sec:propagation-metrics}; boundary ties are handled as specified above.

\begin{table*}[!tp]
\centering
\setlength{\tabcolsep}{4.3pt}
\renewcommand{\arraystretch}{1.10}
\small
\begin{tabular}{llrrrrrrrrr}
\toprule
Benchmark & Method & @10 & @20 & @30 & @40 & @50 & @60 & @70 & @80 & @90 \\
\midrule
BoolQ & AUQ
& 77.70 & 79.89 & 81.64 & 82.99 & 84.89 & 87.73 & 92.46 & 94.71 & 94.29 \\
 & LProp
& \underline{86.78} & \underline{89.50} & \underline{91.86}
& \underline{93.50} & \underline{94.60} & \underline{96.25}
& \underline{98.33} & \underline{99.00} & \underline{100.00} \\
\addlinespace[2pt]
StrategyQA & AUQ
& 70.52 & 72.53 & 71.95 & 72.72 & 72.01 & 72.53 & 73.64 & 75.87 & 76.06 \\
 & LProp
& \underline{73.84} & \underline{76.82} & \underline{79.46}
& \underline{80.56} & \underline{84.17} & \underline{86.46}
& \underline{88.89} & \underline{92.71} & \underline{91.67} \\
\addlinespace[2pt]
MMLU math & AUQ
& 92.11 & 93.39 & 94.09 & 93.67 & 93.26 & 92.66 & 91.66 & 89.62 & 83.73 \\
 & LProp
& \underline{93.99} & \underline{94.84} & \underline{95.00}
& \underline{95.77} & \underline{98.09} & \underline{98.41}
& \underline{97.89} & \underline{98.41} & \underline{96.88} \\
\addlinespace[2pt]
HotpotQA & AUQ
& 56.07 & 57.61 & 57.60 & 58.66 & 60.17 & 62.39 & 66.09 & 68.46 & 68.46 \\
 & LProp
& \underline{53.58} & \underline{58.29} & \underline{61.56}
& \underline{64.61} & \underline{66.02} & \underline{67.80}
& \underline{70.78} & \underline{74.76} & \underline{78.85} \\
\bottomrule
\end{tabular}
\caption{\textbf{Qwen3-8B rejection-rate results.} Retained-answer correctness (\%) at each rejection rate. LProp values are underlined. HotpotQA uses exact match.}
\label{tab:rejection-qwen}
\end{table*}


\section{Inference Token Cost Details}
\label{sec:cost-details}

\begin{table*}[!tp]
\centering
\small
\setlength{\tabcolsep}{4pt}
\resizebox{\textwidth}{!}{%
\begin{tabular}{lrrrrrrrr}
\toprule
& \multicolumn{4}{c}{Qwen3-8B}
& \multicolumn{4}{c}{Mistral-7B-Instruct-v0.3} \\
\cmidrule(lr){2-5}\cmidrule(lr){6-9}
Method & BoolQ & StrategyQA & MMLU math & HotpotQA
& BoolQ & StrategyQA & MMLU math & HotpotQA \\
\midrule
SAUP
& 1,633.0 & 4,071.1 & 11,681.3 & 10,734.7
& 1,824.7 & 4,539.8 & 9,026.2 & 12,710.5 \\
TRACER
& 1,747.3 & 4,184.5 & 11,753.1 & 10,868.1
& 1,953.3 & 4,673.2 & 9,223.6 & 12,765.5 \\
AUQ
& 4,732.7 & 15,833.3 & 10,855.1 & 13,919.3
& 4,487.5 & 12,253.7 & 15,827.4 & 15,564.7 \\
UProp
& 4,489.1 & 9,472.8 & 21,313.3 & 19,002.4
& 5,278.2 & 11,260.5 & 19,150.3 & 23,548.5 \\
\textbf{LProp}
& \textbf{577.7} & \textbf{1,930.9} & \textbf{8,976.7} & \textbf{8,260.3}
& \textbf{597.9} & \textbf{2,119.7} & \textbf{6,582.5} & \textbf{11,222.7} \\
\bottomrule
\end{tabular}}
\caption{\textbf{Mean inference tokens per question ($\downarrow$).} Includes generation input/output, backbone encoding, and auxiliary scoring; excludes training and padding. Bold indicates the lowest value within each setting.}
\label{tab:token-cost}
\end{table*}

\subsection{Token Accounting}

Table~\ref{tab:token-cost} reports mean inference-token positions per question. The total sums generation inputs and outputs, non-generative backbone encoding inputs, and required auxiliary-scoring inputs. It includes source generation, method-specific samples and reflections, and unsuccessful attempts. Repeated context is counted on each generation call.

Reusable encodings are counted once within each method and question; LProp's three heads share these encodings. No cache discount is transferred across methods or questions. Training, development, padding, and counterfactual evaluation probes are excluded. For LProp, source construction and public-evidence encoding remain part of the accounting even though the learned propagation update requests no additional language tokens.

\subsection{Additional Cost Comparisons}

For each dataset--backbone setting, the percentage reduction divides the saved tokens by the comparator's token count. We then average these reductions with equal weight across the eight settings. Relative to AUQ, LProp reduces token usage by 17.3--87.8\%, with an unweighted mean reduction of 61.2\%. Relative to UProp, the reductions range from 52.3\% to 88.7\%, with an unweighted mean of 71.1\%.

On BoolQ/Qwen3, LProp uses 577.7 tokens per question, compared with 4,732.7 for AUQ and 4,489.1 for UProp, while achieving higher F1, AUROC, and AUARC than either method. On MMLU mathematics/Mistral, LProp reduces tokens by 58.4\% relative to AUQ and 65.6\% relative to UProp, alongside the improvements in answer quality and selective reliability reported in the main text.

\subsection{Performance and Token Efficiency}

The cost advantage accompanies higher F1 in all eight settings and higher selective-correctness scores in all eight. On HotpotQA/Qwen3, LProp uses 40.7\% fewer tokens than AUQ and achieves higher AUARC.

As described in Section~\ref{sec:cost-analysis}, source construction supplies the inherited candidates and backbone encoding supplies the public evidence. The learned update revises candidate support directly, without additional textual reflection or trajectory sampling at each handoff. These comparisons quantify inference-token efficiency under the accounting protocol above.


\section{An Empirical Case Study}
\label{app:illustrative-cases}

The recorded episode below shows how LProp revises an inherited error as public reasoning and checking arrive. This single empirical case illustrates both the agent setting and the resulting propagation behavior.

\subsection{A Recorded LProp Recovery Episode}
\label{app:case-evidence-direction}
\label{app:case-complete-handoff}
\label{app:case-lprop-recovery}

We use a recorded Qwen3-8B episode to connect the multi-step construction in Appendix~\ref{sec:multistep-agent-protocol} with an observed answer recovery. The question comes from MMLU high-school statistics. A private solver produces the source draft; a public solver receives the question and options; a checker subsequently reviews the public derivation. These roles execute in the prescribed order. The receiver obtains the source carrier and successive public encodings, with private draft content retained through the carrier. The probabilities below are LProp's answer distributions at the corresponding stages, rather than the public solver's confidence scores.

\begin{figure}[!tbp]
\begin{uncertaintycase}{green}{Case Study: Recovery through Propagation}
\small
\textbf{Question.} What does the difference between the 60th and 40th percentile scores measure?

\textbf{Options.} A: central tendency; \textbf{B: variability}; C: symmetry; D: skewness.
\par\smallskip
\begingroup
\pgfmathsetlengthmacro{\casewidth}{\linewidth-25pt}
\begin{tikzpicture}[
  stage/.style={
    draw=black!45,
    rounded corners=2pt,
    fill=white,
    text width=\casewidth,
    inner sep=3pt,
    align=left,
    font=\fontsize{9}{10.5}\selectfont,
    anchor=north west
  },
  flow/.style={
    -{Latex[length=4pt,width=3pt]},
    line width=.65pt
  },
  flowlabel/.style={
    font=\fontsize{8.5}{10}\selectfont,
    fill=green!4!white,
    inner sep=1pt
  }
]
\node[stage] (source) at (0,0) {
  \textbf{1. Private draft: D (incorrect)}\\[1pt]
  Interprets the percentile gap as skewness.\\[1pt]
  \textbf{D: 91.56\%}\quad B: 5.11\%\\
  Selected-answer uncertainty: 8.44\%
};

\node[
  stage,
  below=25pt of source.south west,
  anchor=north west
] (derive) {
  \textbf{2. After public derivation: B (correct)}\\[1pt]
  Explains that the gap measures spread.\\[1pt]
  D: 21.72\%\quad
  \textbf{\textcolor{blue!55!black}{B: 76.93\%}}\\
  Selected-answer uncertainty: 23.07\%
};

\node[
  stage,
  below=16pt of derive.south west,
  anchor=north west
] (check) {
  \textbf{3. After public check: B (correct)}\\[1pt]
  Confirms that spread means variability.\\[1pt]
  D: 11.23\%\quad
  \textbf{\textcolor{blue!55!black}{B: 87.32\%}}\\
  Selected-answer uncertainty: 12.68\%
};

\node[
  stage,
  fill=black!4,
  below=8pt of check.south west,
  anchor=north west
] (frozen) {
  \textbf{Same model, updates disabled}\\[1pt]
  Source D \,$\longrightarrow$\, Final D (incorrect)\\
  Support for D stays at \textbf{91.56\%}.
};

\draw[flow]
  (source.south) --
  node[flowlabel,right=2pt,align=left]
  {Latent handoff\\+ evidence update}
  (derive.north);

\draw[flow]
  (derive.south) --
  node[flowlabel,right=2pt]
  {Evidence update}
  (check.north);

\draw[flow,dashed,black!65]
  (source.east) -- ++(12pt,0) |- (frozen.east);

\end{tikzpicture}
\endgroup
\par\smallskip
\textit{Qwen3-8B. Solid arrows: LProp updates; dashed branch: frozen-update control. Texts are condensed from the recorded episode.}
\end{uncertaintycase}
\end{figure}

\begin{table*}[!tp]
\centering
\setlength{\tabcolsep}{3.3pt}
\renewcommand{\arraystretch}{1.04}
\fontsize{8.2}{9.6}\selectfont
\begin{tabular}{lrrrrrrrrrrrr}
\toprule
& \multicolumn{3}{c}{BoolQ}
& \multicolumn{3}{c}{StrategyQA}
& \multicolumn{3}{c}{MMLU mathematics}
& \multicolumn{3}{c}{HotpotQA} \\
\cmidrule(lr){2-4}
\cmidrule(lr){5-7}
\cmidrule(lr){8-10}
\cmidrule(lr){11-13}
Initial $W/C$
& \multicolumn{3}{c}{291 / 709}
& \multicolumn{3}{c}{167 / 313}
& \multicolumn{3}{c}{33 / 281}
& \multicolumn{3}{c}{267 / 245} \\
$N$
& \multicolumn{3}{c}{1,000}
& \multicolumn{3}{c}{480}
& \multicolumn{3}{c}{314}
& \multicolumn{3}{c}{512} \\
\midrule
Receiver
& $W\!\to\!C$ & $C\!\to\!W$ & Final $C$
& $W\!\to\!C$ & $C\!\to\!W$ & Final $C$
& $W\!\to\!C$ & $C\!\to\!W$ & Final $C$
& $W\!\to\!C$ & $C\!\to\!W$ & Final $C$ \\
\midrule
Full LProp
& \underline{182} & \underline{51} & \AppendixBest{\underline{840}}
& \underline{73} & \underline{25} & \AppendixBest{\underline{361}}
& \underline{13} & \underline{7} & \AppendixBest{\underline{287}}
& \underline{21} & \underline{3} & \AppendixBest{\underline{263}} \\
\addlinespace[2pt]
\multicolumn{13}{l}{\textit{Handoff representation}} \\
Probability-only
& 144 & 72 & 781
& 33 & 26 & 320
& 8 & 7 & 282
& 1 & 4 & 242 \\
Text summary
& 169 & 53 & \AppendixSecond{825}
& 63 & 24 & \AppendixSecond{352}
& 12 & 8 & \AppendixSecond{285}
& 11 & 2 & \AppendixSecond{254} \\
\addlinespace[2pt]
\multicolumn{13}{l}{\textit{Latent content}} \\
Frozen content
& 165 & 91 & 783
& 76 & 61 & 328
& 7 & 7 & 281
& 3 & 1 & 247 \\
Gaussian noise
& 130 & 92 & 747
& 21 & 55 & 279
& 10 & 33 & 258
& 0 & 11 & 234 \\
\addlinespace[2pt]
\multicolumn{13}{l}{\textit{Receiver design}} \\
Matched MLP
& 153 & 43 & 819
& 55 & 34 & 334
& 8 & 13 & 276
& 6 & 4 & 247 \\
Log-linear fusion
& 95 & 38 & 766
& 19 & 16 & 316
& 5 & 3 & 283
& 1 & 1 & 245 \\
\bottomrule
\end{tabular}
\caption{\textbf{Matched receiver ablation counts with frozen Qwen3-8B.} $W$ and $C$ denote wrong and correct answers. The two transition columns count recovered errors and introduced errors. Gray shades mark the best and second-best final-correct counts within each benchmark; full-model values are underlined.}
\label{tab:receiver-ablation}
\end{table*}

\paragraph{How uncertainty changes during recovery.}
\looseness=-1
\textbf{LProp revises both the inherited answer and its uncertainty as evidence arrives.} The derivation transfers support from D to the retained alternative B; the subsequent check strengthens B. Selected-answer uncertainty first rises when the initial confidence is revised, then falls after confirmation. Freezing the update preserves the initial error under the same source state and answer-selection rule. The public derivation and check follow the fixed evidence schedule; their execution is not triggered by the displayed uncertainty values. This control isolates enabling the update for this recorded carrier and episode. The separately trained public-evidence-only control in Appendix~\ref{app:receiver-ablation} tests the contribution of inherited content.

\paragraph{Recorded probability distributions.}
The workflow highlights B and D; the complete distributions are given below. Uncertainty is one minus the selected-answer probability.
\par\smallskip
{\centering
\footnotesize
\setlength{\tabcolsep}{3pt}
\renewcommand{\arraystretch}{1.08}
\begin{tabular}{@{}lrrrrc@{}}
\toprule
Stage & $P(A)$ & $P(B)$ & $P(C)$ & $P(D)$ & Answer \\
\midrule
Source        & .0065 & .0511 & .0268 & .9156 & D \\
Derivation    & .0045 & .7693 & .0090 & .2172 & B \\
Check         & .0069 & .8732 & .0076 & .1123 & B \\
Frozen update & .0065 & .0511 & .0268 & .9156 & D \\
\bottomrule
\end{tabular}\par}

\paragraph{Case selection.}
\looseness=-1
This qualitative example is selected retrospectively from the evaluation records. Values average the development-selected heads with seeds 17, 29, and 43. The frozen-update control uses the same model with public updates disabled.


\section{Matched Receiver Ablations}
\label{app:receiver-ablation}

\subsection[Do latent content and its update improve recovery?]{
  \AppendixRQBadge{Q2}\enspace
  Do latent content and its update improve recovery?
}
\label{app:receiver-research-question}

\paragraph{Question and rationale.}
\looseness=-1
We test whether recovery depends on the handoff representation, informative latent content and its updating, or the receiver design.

\paragraph{Setting.}
\looseness=-1
We freeze Qwen3-8B and compare seven receiver variants using the main experiment's splits. Questions, source-private evidence, candidates, supervision, and public evidence remain fixed, including each benchmark's multi-step schedule (Appendix~\ref{sec:multistep-agent-protocol}). Full LProp is the retrained reference and receives both source probabilities $w$ and latent content $h$.

\paragraph{Controls.}
\looseness=-1
We design different control type:

\textbf{Handoff representation:} probability-only retains $w$ and zeros source $h$; its BoolQ receiver can still construct states from public evidence. Text summary transfers a 128-token summary of the private evidence.

\textbf{Latent content:} frozen content disables revision of source $h$ while allowing probability updates; Gaussian noise replaces source $h$ with a Gaussian-noise carrier.

\textbf{Receiver design:} matched MLP uses source probabilities and public-evidence features without source $h$, with parameters within 1\% of the full receiver; log-linear fusion combines source and public probabilities using two learned coefficients. The probability-only, frozen-content, matched-MLP, and fusion controls are retrained for their respective interfaces.

\paragraph{Metrics.}
\looseness=-1
Figure~\ref{fig:receiver-ablation} shows recovery $100R/N$, regression $-100D/N$, and net gain $100(R-D)/N$, where $N$ is the number of questions. Correctness follows the main protocol, including HotpotQA exact match. Table~\ref{tab:receiver-ablation} reports counts; final correct equals initial correct plus recovered minus regressed.

\paragraph{Results.}
\looseness=-1
Full LProp has the highest final correctness in every setting. It recovers more errors than probability-only and text-summary handoffs. Gaussian noise reduces final correctness across all benchmarks; on StrategyQA, frozen content recovers slightly more errors but introduces substantially more regressions. Full LProp also outperforms matched MLP and log-linear fusion, supporting the complete receiver design.

\section{Additional Related Work}
\label{app:additional-related-work}

Large language models (LLMs) have advanced rapidly in recent years \citep{ye2025llms4all, xiong2026adaptive, xiong2026dive}. Building on this progress, LLM-driven agents and agentic reinforcement learning have gained increasing attention for their ability to plan, interact with external environments, and solve complex tasks with limited human oversight \citep{zhang2026optimizing, dong2025agentic, dong2026agentic}. In this section, we provide a boarder review of our related work. 

\paragraph{Uncertainty, interpretation, and commitment.}
Measuring disagreement among possible answers and deciding what to communicate are distinct reliability problems. Kernel Language Entropy addresses the former by replacing hard semantic clusters with a positive-semidefinite kernel that captures graded similarities and by computing its von Neumann entropy~\citep{NEURIPS2024_10c456d2}. SAGE uses answer-conditioned targets to calibrate verbal expressions of uncertainty, while GCPO combines geometry-aware disagreement with reward-based calibration to guide policy optimization~\citep{shi2026sage,zhang2026semantic}. Linguistic calibration places the target at the reader, training the model to generate responses that improve the reader's probabilistic forecasts~\citep{pmlr-v235-band24a}. Conformal methods use calibration to control what the system returns. Conformal generation uses sampling and rejection to construct candidate sets with a coverage guarantee for an acceptable response~\citep{quach2024conformal}. Conformal factuality controls the risk of unsupported assertions by backing off to less specific claims~\citep{pmlr-v235-mohri24a}. These guarantees depend on the calibration data and the distributional assumptions of each method.

\paragraph{Detecting and filling information gaps.}
The signal that flags a possible information gap and the source that can fill it play different roles. SelfCheckGPT compares sampled responses to flag potentially inconsistent factual statements~\citep{manakul-etal-2023-selfcheckgpt}. When external evidence can address a gap, FLARE uses low-confidence predicted tokens to trigger retrieval~\citep{jiang-etal-2023-active}, while Self-RAG learns when to retrieve and uses reflection tokens to critique retrieved passages and generated text~\citep{ICLR2024_25f7be96}. Once evidence has been retrieved, CRAG evaluates it to choose among refinement, web search, and a combination of the two~\citep{yan2024corrective}. CiteAudit adds source verification to the evidence pipeline~\citep{shi2026citeaudit}. Clarification addresses gaps in what the user has communicated. Drift-Bench studies clarification across input faults and user personas~\citep{bao2026drift, zhang2026beyond}, while masked-user-reply experiments motivate assigning training credit based on representations of unmet requirements~\citep{bao2026not}.

\paragraph{Allocating computation and context.}
The timing of an allocation decision determines which signals it can use. Before generation, HybridLLM predicts query difficulty to balance quality and cost~\citep{ding2024hybridllm}, while RouteLLM routes queries using learned model preferences~\citep{ICLR2025_5503a7c6}. Once answers are available, FrugalGPT scores completed responses and escalates through a model cascade when scores fall below optimized thresholds~\citep{chen2024frugalgpt}, while EvolveRouter uses answer agreement to decide when to stop~\citep{huang2026evolverouter}. Allocation also extends to the agents, context, and evidence used to solve a task. AgentRouter and NG-Router organize selection around graph-structured agent--context relations and evidence subgraphs, respectively~\citep{zhang2026agentrouter,shi2026ng}. Across repeated tasks, LLMA-Mem shows that reusing experience can let smaller teams outperform larger ones~\citep{wu2026scaling}.

\paragraph{Preserving information for revision.}
What a downstream stage can revise depends on the information it receives. RARR takes a completed generation, retrieves evidence, and edits unsupported passages while seeking to preserve the rest~\citep{gao-etal-2023-rarr}. Multiagent debate revises proposed answers through successive rounds of exchanged answers and reasoning, with agreement emerging from deliberation among model instances~\citep{pmlr-v235-du24e}. Both approaches operate on generated text. LProp studies what candidate-specific information a handoff must preserve so that new evidence can update both the latent content of individual candidates and the probability mass assigned to them.
\end{document}